\documentclass[openacc]{rstransa}

\usepackage{graphicx}
\usepackage{subcaption}
\usepackage{booktabs,threeparttable,makecell,array}
\usepackage{ragged2e} 
\usepackage{multirow}

\titlehead{Research}

\begin{document}

\title{Limitations of Public Chest Radiography Datasets for Artificial Intelligence: Label Quality, Domain Shift, Bias and Evaluation Challenges}

\author{
A. Rafferty$^{1}$ and A. Rajan$^{1}$}

\address{$^{1}$School of Informatics, University of Edinburgh, UK\\
}

\subject{Artificial Intelligence, Medical Imaging, Public Datasets}


\corres{Amy Rafferty\\
\email{s1817812@ed.ac.uk}}

\begin{abstract}
Artificial intelligence has shown significant promise in chest radiography, where deep learning models can approach radiologist-level diagnostic performance. Progress has been accelerated by large public datasets such as MIMIC-CXR, ChestX-ray14, PadChest, and CheXpert, which provide hundreds of thousands of labelled images with pathology annotations. However, these datasets also present important limitations. Automated label extraction from radiology reports introduces errors, particularly in handling uncertainty and negation, and radiologist review frequently disagrees with assigned labels. In addition, domain shift and population bias restrict model generalisability, while evaluation practices often overlook clinically meaningful measures. We conduct a systematic analysis of these challenges, focusing on label quality, dataset bias, and domain shift. Our cross-dataset domain shift evaluation across multiple model architectures revealed substantial external performance degradation, with pronounced reductions in AUPRC and F1 scores relative to internal testing. To assess dataset bias, we trained a source-classification model that distinguished datasets with near-perfect accuracy, and performed subgroup analyses showing reduced performance for minority age and sex groups. Finally, expert review by two board-certified radiologists identified significant disagreement with public dataset labels. Our findings highlight important clinical weaknesses of current benchmarks and emphasise the need for clinician-validated datasets and fairer evaluation frameworks.
\end{abstract}

\begin{fmtext}
\section{Introduction}

Artificial intelligence is increasingly being adopted in

\end{fmtext}
\maketitle

healthcare domains, contributing to key applications like automated image interpretation, disease detection, clinical triage, workflow optimisation, and patient risk stratification \cite{assist1, assist2}. In diagnostic radiology, deep learning models have demonstrated high accuracy in identifying conditions such as pneumonia, pneumothorax, and pulmonary nodules on chest radiographs—approaching or even exceeding the performance of experienced radiologists \cite{comparable1, comparable3}. The integration of AI tools into healthcare holds significant promise for improving diagnostic efficiency, alleviating workforce constraints, and extending access to expert-level interpretation across diverse clinical settings \cite{benefit}.
The research and growth of medical AI has been supported by the availability of large, publicly accessible imaging datasets that enable reproducible algorithm development and rigorous benchmarking. In chest radiography, there is a wide range of datasets available \cite{datasets}, ranging from small collections of under 1,000 images (e.g., Montgomery \cite{montgomery}, COVID-CXR \cite{covid-cxr}) to large collections of over 100,000 images such as MIMIC-CXR \cite{mimic, mimic-jpg, physionet}, ChestX-ray14 \cite{cxr8}, PadChest \cite{padchest} and CheXpert \cite{chexpert, chexpert-plus}. These large public datasets remain the basis for much of current chest imaging AI research \cite{datasets}.

\begin{figure}[h]
    \centering
    \vspace{-8pt}
    \includegraphics[width=0.6\linewidth]{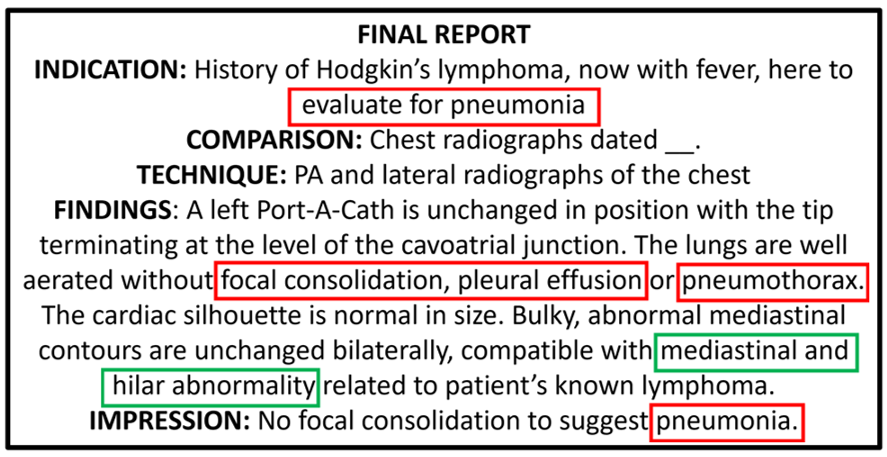}
    \caption{Example radiography report from the MIMIC-CXR dataset. The associated chest radiograph is labelled as \textbf{Lung Cancer} and \textbf{Pneumonia} by the automated CheXpert labeller. \textbf{Pneumonia} indicators (red) are negative mentions, leading to a false diagnosis.}
    \label{fig:report}
    \vspace{-8pt}
\end{figure}

While these resources enable large-scale model development and benchmarking, they are also known to exhibit critical challenges that complicate clinical translation. The most notable problems are imperfect labels derived from automated text mining approaches \cite{labelbad, chexpertplusplus}, demographic and institutional biases that induce domain shift \cite{bias, bias2, bias3}, and evaluation protocols that do not reliably reflect clinical operating points \cite{eval, eval2}. Collectively, these factors can produce deceptively high internal performance that does not persist under external validation \cite{chexbert}. 
Automated report mining has enabled dataset curation at scale for datasets such as CheXpert, MIMIC-CXR, ChestX-ray14, and PadChest, but remains vulnerable to negation, uncertainty, and study-level misalignment, leading to false positives/negatives and label drift across conditions and sites \cite{unreliable2}. Such noise is sufficient for training, yet it constrains model reliability when radiologist consensus is the standard. An example case of automated labelling leading to a false positive for Pneumonia is shown in Figure \ref{fig:report}.
Most large public datasets originate from single centres or health systems and differ in populations, acquisition hardware, protocols, and preprocessing. Models can therefore exploit site-specific artefacts (e.g., devices, overlays, or processing fingerprints) rather than pathologies, yielding performance that collapses when scanners or workflows change \cite{shortcut, shortcut2}.
While AUROC and related metrics are useful and widely adopted for ranking classifiers, they obscure clinically meaningful operating points and the asymmetric costs of false positives versus false negatives \cite{eval, eval2}. Under class imbalance, AUPRC and threshold-dependent metrics (Sensitivity, Specificity, F1) are more informative \cite{auroc}, but only when decision thresholds are specified and validated appropriately. 
In this study, we make the following contributions:

\vspace{-10pt}

\begin{itemize}
    \item We provide a detailed overview of the critical limitations of public chest radiography datasets, specifically label quality, domain shift and dataset bias, and discuss their impact and potential solutions.
    \item We conduct a cross-dataset empirical study of domain shift by training seven widely used deep learning model architectures on each of four large public datasets, and evaluating performance across a range of metrics for internal and external test sets. We provide a comprehensive performance comparison across datasets, using both threshold-independent (AUROC, AUPRC) and threshold-dependent (Specificity, Sensitivity, F1) metrics. We visualise and quantify significant degradation from internal to external evaluation, identifying the source–target pairs most affected by domain shift.
    \item We analyse dataset bias and generalisability by combining cross-dataset evaluation with (i) a dataset-source classification diagnostic that reveals strong non-clinical dataset signatures, and (ii) subgroup analyses by age and sex under fixed clinical operating thresholds, demonstrating systematic disparities for underrepresented cohorts.
    \item We perform an expert audit in which two board-certified radiologists review ground-truth labels for a subset of MIMIC-CXR and CheXpert images via agree/disagree judgements, revealing substantial disagreement with the public labels.
\end{itemize}

\vspace{-10pt}
\section{Challenges in Public Chest Radiography Datasets}

Public chest radiography datasets have become essential resources for developing and benchmarking AI models, but they suffer from important limitations. In this section, we outline four major dataset challenges: label quality and semantic noise introduced by automated annotations, demographic and institutional biases that limit representativeness and create domain shift, technical artefacts and medical devices that promote shortcut learning, and the limitations of evaluation practices. Together, these issues explain why models trained on widely used datasets may achieve high reported accuracy yet fail to generalise reliably in clinical settings.

\vspace{-5pt}
\subsection{Label Quality and Semantic Noise}

One of the most significant challenges across publicly available chest radiography datasets is the quality of diagnostic labels. Large-scale resources such as CheXpert \cite{chexpert} and MIMIC-CXR \cite{mimic} derive their annotations from accompanying radiology reports using rule-based natural language processing (NLP) pipelines. The CheXpert labeller \cite{chexpert}, for example, produces labels for 14 common thoracic findings by parsing free-text radiology reports. Similarly, the NIH ChestX-ray14 dataset \cite{cxr8} used the NegBio system \cite{negbio} to extract labels from unstructured text, while the PadChest \cite{padchest} dataset employed recurrent neural networks to process Spanish-language reports. These automated systems enabled dataset construction at scale, but remain unreliable when confronted with negations (e.g., “no evidence of pneumonia”) and expressions of diagnostic uncertainty (e.g., “cannot exclude effusion”). Also, most labelling pipelines assign labels at the study or report level rather than the image level, creating systematic mismatches between pixel-level evidence and the assigned diagnosis \cite{unreliable2, unreliable3}.
Recent re-labelling efforts highlight the scope of this problem. Contributions such as CheXpert++ \cite{chexpertplusplus} and CheXbert \cite{chexbert} have demonstrated frequent divergence between automated labels and those assigned by board-certified radiologists, revealing substantial false-positive and false-negative rates across key conditions such as atelectasis and consolidation. High-prevalence pathologies like cardiomegaly and pleural effusion are inconsistently annotated \cite{labelbad, unreliable2}. In contrast, some smaller datasets such as JSRT \cite{jsrt} contain exclusively expert-curated annotations, and therefore achieve higher label fidelity but lack the scale needed for training deep learning models, often including fewer than 1,000 images and covering only specific pathologies \cite{montgomery, datasets}.

\vspace{-8pt}
\paragraph{Impact. }
Models trained on incorrect labels can appear to achieve high performance when benchmarked against flawed ground truth, but fail in real-world deployment where radiologist consensus is the standard. Deep learning models are especially prone to overfitting annotation errors, effectively learning to replicate dataset-specific mistakes rather than true pathology recognition \cite{pitfalls}. Such findings highlight how label quality directly constrains model reliability, generalisability, and clinical trustworthiness.

\vspace{-8pt}
\paragraph{Potential Solutions. }
Several strategies may mitigate these issues. Systematic expert validation of representative subsets provides an estimate of mislabelling rates and a gold-standard reference for benchmarking, as shown in CheXpert++ \cite{chexpertplusplus}. While large-scale manual relabelling remains impractical, integrating domain expertise during both NLP pipeline development and dataset evaluation can improve label fidelity. Uncertain findings should also be explicitly retained rather than collapsed into binary classes, preserving the spectrum of diagnostic ambiguity common in clinical radiology. There are currently conflicting methods on how to deal with uncertainties. In CheXpert, uncertain findings are explicitly retained, whereas in ChestX-ray14 they are often collapsed into binary outcomes. Some studies have proposed treating uncertainty as a separate predictive category \cite{uncertain}, while others advocate probabilistic labelling to reflect clinical ambiguity \cite{uncertain2}. There is no consensus, but retaining uncertainty rather than discarding it appears more faithful to clinical practice. Another important factor is transparency - dataset releases should document the specific NLP tool, version, rules, and uncertainty-handling strategies used in annotation. This practice not only facilitates reproducibility but also enables fair comparison across datasets and supports iterative refinement of labelling pipelines over time.

\vspace{-5pt}
\subsection{Dataset Bias, Domain Shift, and Shortcut Learning}

\newcommand{\msff}[2]{\makecell{ #1\\ #2}}

Another major challenge regarding public chest radiography datasets is the presence of demographic, institutional, and acquisition biases that limit generalisability across populations. Most widely used datasets originate from a single institution or health system, leading to sampling that is not representative of broader clinical settings. For example, MIMIC-CXR was collected exclusively at Beth Israel Deaconess Medical Center in Boston, USA \cite{mimic}, while CheXpert was derived entirely from Stanford Hospital, USA \cite{chexpert}. Similarly, ChestX-ray14 was built from a single NIH Clinical Center repository \cite{cxr8}, and PadChest, though larger and more geographically distinct, remains confined to a Spanish cohort \cite{padchest}. This lack of geographic and institutional diversity limits external validity, particularly given known global heterogeneity in disease prevalence, imaging protocols, and patient populations \cite{datasets}.
Multiple studies have shown that public chest radiography datasets underrepresent paediatric, elderly, and minority populations \cite{chexclusion, bias}. MIMIC-CXR, for example, overrepresents critically ill ICU patients, producing label distributions that diverge substantially from general hospital populations \cite{mimic}. In ChestX-ray14, only adult patients are included, omitting paediatric findings entirely \cite{cxr8}. Smaller datasets such as Montgomery \cite{montgomery} and Shenzhen Hospital TB Set \cite{montgomery} were specifically curated for tuberculosis screening, providing narrow diagnostic coverage that does not reflect broader clinical practice. These sampling constraints bias models towards particular pathologies and patient groups, increasing the risk of unfair or inaccurate predictions when deployed in diverse populations.

Domain shift further arises from differences in acquisition hardware, image preprocessing, and institutional protocols. Variability in detectors, exposure settings, positioning, and preprocessing pipelines introduces systematic shifts that deep learning models can exploit as spurious cues \cite{bias2, bias3}. Models trained on public datasets sometimes achieve high performance by learning dataset-specific artefacts, such as embedded text markers or laterality labels in CheXpert and MIMIC-CXR \cite{mimic, chexpert}, rather than physiological features \cite{pitfalls, signal}. Medical devices represent another problem. In ICU-focused datasets such as MIMIC-CXR, endotracheal tubes, chest drains, and ventilator equipment are frequently visible, providing indirect signals of disease severity. Models may therefore learn to associate these devices with particular diagnoses, rather than identifying visual features of lung pathologies \cite{signal, shortcut}. Similar issues have been noted in CheXpert, where pacemakers and defibrillators inadvertently serve as proxies for chronic cardiac disease \cite{pitfalls}. 

A concrete diagnostic of dataset bias is the extent to which a model can infer dataset identity from images alone. Classic "name-that-dataset" analyses show that strong dataset classifiers can be trained by exploiting non-pathological signatures, and that such signatures hinder cross-dataset generalisation \cite{bias_rebut, bias_rebut2}. This phenomenon has also been observed in medical imaging, including chest radiography benchmarks, where acquisition and processing "fingerprints" may be readily learnable \cite{bias_rebut3, bias_rebut4}. 

Recent work has highlighted the importance of identifying dataset shift prior to clinical deployment, including methods that explicitly detect distributional differences between training and target datasets \cite{automatic_dataset_shift}. While such approaches aim to flag the presence of shift, our focus is complementary: we quantify the downstream impact of domain shift on diagnostic performance across multiple architectures and institutions. By directly evaluating cross-dataset generalisation, we characterise how dataset shift manifests as systematic performance degradation under realistic deployment conditions.

Pre-processing pipelines contribute further to domain shift. Public datasets undergo largely different downsampling and contrast normalisation processes prior to release, creating characteristic intensity patterns \cite{shortcut, shortcut2}. Most datasets also contain images of greatly reduced resolutions compared to clinical DICOM standards \cite{datasets}. While this downsampling facilitates practical storage and distribution, it reduces image quality and can obscure small pathological indicators, such as pulmonary nodules and fine interstitial markings. Higher-resolution inputs have been shown to improve classification performance and enable more reliable detection of subtle pathologies using chest radiography datasets \cite{resolution}. The fact that most public datasets do not document acquisition parameters, exposure settings, or pre-processing decisions further compounds these issues, making it difficult to identify or mitigate technical biases.

\vspace{-8pt}
\paragraph{Impact. }
These factors mean that models may achieve deceptively high internal accuracy by exploiting dataset-specific artefacts or preprocessing fingerprints, resulting in inflated performance estimates that collapse under external validation \cite{zech, shortcut, shortcut2}. Subtle dependencies on non-generalising cues can destabilise models, as performance deteriorates when scanners, protocols, or patient distributions differ. This lack of robustness undermines clinical translation, as deployment environments rarely match the narrow sampling distributions of these datasets.

\vspace{-8pt}
\paragraph{Potential Solutions. }
Addressing these challenges requires coordinated methodological and infrastructural solutions. Cross-institutional benchmarking should be standard practice, ensuring that models are validated across datasets with different demographic and technical characteristics. Dataset curators should provide detailed metadata on scanner types, acquisition protocols, known artefacts and pre-processing decisions, enabling researchers to stratify analyses and identify hidden biases. Rigorous artefact auditing should be standard practice, with systematic screening for non-anatomical content and spurious features. Frameworks that evaluate subgroup performance (e.g., sex, age) are essential for exposing disparities prior to deployment. The creation of multi-centre, internationally representative datasets represents the most powerful strategy for mitigating bias and improving generalisability, though this remains infeasible.

\vspace{-5pt}
\subsection{Evaluation Practices and Metrics}

Even when trained on large public chest radiography datasets, the evaluation of deep learning models often relies on metrics and validation strategies that do not align with clinical practice. The vast majority of published studies report model performance in terms of classification precision, recall, and the area under the receiver operating characteristic curve (AUROC). While these metrics are informative for comparing classifiers, they fail to capture clinically relevant outcomes, such as decision thresholds, and the cost of false positives and false negatives \cite{eval, eval2}. In a medical context, the consequences of an error are highly asymmetric: a false negative in pneumonia detection may delay life-saving treatment, whereas a false positive may lead to unnecessary antibiotics and increased healthcare costs \cite{eval}. 
Another problem is that the classification performance of deep learning models is often exclusively evaluated against the automated labels associated with the public datasets they are trained on. As discussed earlier, these labels, gathered through automated NLP techniques such as the CheXpert labeller \cite{chexpert} or NegBio \cite{negbio}, suffer from substantial error rates. 

\vspace{-8pt}
\paragraph{Impact. }
When models are evaluated against flawed labels, they risk learning and reproducing annotation errors rather than detecting true radiological findings. Small-scale label evaluations conducted by medical experts, such as those performed in CheXpert++ \cite{chexpertplusplus}, have demonstrated frequent divergence between automated and radiologist labels, raising concerns about the reliability of published benchmarking results. Notably, there is minimal integration of human expertise in evaluation pipelines. Most model assessments are conducted purely against computational metrics, with little to no involvement of radiologists in verifying clinical plausibility. Without expert evaluation, it is impossible to determine whether model predictions are aligned with real-world diagnostic reasoning.

\vspace{-8pt}
\paragraph{Potential Solutions. }
Several strategies could strengthen evaluation practices. Metrics should extend beyond standard computational metrics to include clinically grounded measures of diagnostic utility. Expert radiologist involvement is essential not only for label refinement but also for assessing model predictions in clinically relevant contexts. A more rigorous, clinician-centred evaluation framework will better capture the reliability and utility of chest radiography AI systems.

\vspace{-5pt}
\section{Empirical Analysis: Domain Shift}

To assess domain shift, we adopt a cross-dataset evaluation framework spanning four large-scale public chest radiography datasets: MIMIC-CXR, CheXpert, PadChest, and ChestX-ray14. Each dataset is treated in turn as the source: models are trained using its training split, operating thresholds are calibrated on the corresponding validation split, and performance is assessed both on the source test split (in-distribution) and on the test splits of the remaining datasets (external). By comparing internal and external performance, we aim to quantify the extent to which models trained on widely used public datasets degrade when applied to novel data distributions.

\vspace{-5pt}
\subsection{Datasets}\label{data}

We use four large-scale, publicly available chest radiography datasets.
\textbf{MIMIC-CXR} contains 377,110 chest radiographs from 227,835 studies at the Beth Israel Deaconess Medical Center (Boston, USA), specifically ICU data, collected between 2011 and 2016. Each image is paired with an anonymised free-text radiology report, and labelled automatically through natural language processing using the CheXpert labeller.
\textbf{CheXpert} contains 224,316 chest radiographs from Stanford Hospital (USA), collected between October 2002 and July 2017, including both inpatient and outpatient exams. Each image is again paired with an anonymised free-text radiology report, and labelled automatically through natural language processing using the CheXpert labeller.
\textbf{NIH ChestX-ray14} contains 112,120 frontal chest radiographs from the NIH Clinical Center (Maryland, USA), annotated with 14 binary disease labels via natural language processing. Reports and NLP algorithms are not publicly available.
\textbf{PadChest} contains 160,868 chest radiographs from multiple viewpoints, collected from Hospital San Juan (Spain) between 2009 and 2017. Images are paired with multi-label pathology annotations based on translated Spanish radiological reports, 27\% of which were manually reviewed.

These datasets form the backbone of most recent studies analysing deep learning models for chest radiography, and are frequently used for both model training and benchmarking. Within these datasets we concentrate on six clinically significant findings: \textbf{No Finding}, \textbf{Cardiomegaly}, \textbf{Lung Cancer}, \textbf{Pneumonia}, \textbf{Pneumothorax} and \textbf{Pleural Effusion}.
To control for confounding technical variables, only posterior–anterior (PA) view chest radiographs were included, as view position is a well-documented source of domain shift \cite{pa_usage,pa_usage2}. To address class imbalance, we applied one-sided selection undersampling \cite{onesidedselection}. Data is then split into 80/10/10 train/validation/test partitions on a patient basis to ensure no subject appeared in more than one split. Table~\ref{tab:classdist} summarises the resulting dataset sizes and class distributions.

\begin{table}[t]
\caption{Summary of dataset sizes and class distributions after pre-processing.}
\label{tab:classdist}
\small
\vspace{-5pt}
\begin{tabular}{lllll}
\hline
\textbf{Label} & \textbf{MIMIC-CXR} & \textbf{CheXpert} & \textbf{ChestX-ray14} & \textbf{PadChest} \\
\hline
No Finding & 16203 & 3195 & 21758 & 15643 \\
Pneumonia & 3228 & 1198 & 594 & 547 \\
Pneumothorax & 1867 & 1462 & 1556 & 48 \\
Cardiomegaly & 5737 & 2909 & 710 & 2745 \\
Lung Cancer & 1429 & 2122 & 3507 & 1229 \\
Pleural Effusion & 7428 & 3294 & 3211 & 605 \\
\hline
Total & 35892 & 14180 & 31419 & 20817 \\
\hline
\end{tabular}
\vspace{-5pt}
\end{table}

\vspace{-5pt}
\subsection{Model Architectures}
We select backbone model architectures used by the best-performing submissions in the recent MICCAI CXR-LT challenge \cite{cxr-lt, physionet}, which emphasised imbalanced classification on MIMIC-CXR. The chosen models are ResNet50 \cite{resnet}, ResNet101 \cite{resnet}, ResNeXt101 \cite{resnext}, DenseNet161 \cite{dense}, ConvNeXt-S \cite{convs}, ConvNeXt-B \cite{convs}, and EfficientNetV2-S \cite{efficient}. This selection represents a diverse set of convolutional networks that have achieved state-of-the-art results in chest radiography tasks, and evaluating them directly allows us to test whether architecture choice impacts domain generalisability. 
All models were implemented in Pytorch and trained for up to 50 epochs with early stopping enabled. We use stochastic gradient descent (learning rate = 0.001, momentum = 0.9, weight decay = 0) with a batch size of 8 and cross-entropy loss. A fixed random seed of 34 was used for all runs. Training data was shuffled at each epoch, and model checkpoints were selected based on the minimum validation loss.


\vspace{-5pt}
\subsection{Label Harmonisation}
\label{sec:labelharm}

We explicitly document below the label-mapping rules used to harmonise pathology categories across datasets, including the clinical rationale for each label choice and the exclusion of ambiguous or non-specific labels to ensure semantic consistency.

For cross-dataset analysis, we focus on six clinically significant and commonly reported chest radiography findings: \textbf{No Finding, Pneumonia, Pneumothorax, Cardiomegaly, Lung Cancer,} and \textbf{Pleural Effusion}. These were selected based on their frequent presence in imaging interpretation and their critical impact on patient care. For example, pneumonia and lung cancer, specifically lung nodules, are among the most common sources of missed diagnoses on chest radiographs \cite{missed1} and contribute to nearly 43\% of radiography malpractice claims. These findings are also among the most prevalent labels in large-scale public chest radiography datasets \cite{chexpert, mimic, cxr8}.

For CheXpert and MIMIC-CXR, direct label mappings were possible since both datasets use the same label ontology defined by the CheXpert labeller \cite{chexpert}. We note that we take the \textit{Lung Lesion} label from these datasets to indicate \textbf{Lung Cancer}, under guidance of a consultant radiologist. For ChestX-ray14, to achieve label harmonisation we mapped \texttt{Mass} and \texttt{Nodule} into a combined \textbf{Lung Cancer} category, and \texttt{Effusion} to \textbf{Pleural Effusion}, again under guidance. For PadChest, which is a Spanish dataset, we leveraged its built-in concept normalisation framework to map Spanish-language report terms to standardised clinical concepts. This allowed us to identify direct and unambiguous equivalents for the target labels (e.g., \textit{Neumonía} $\rightarrow$ \textbf{Pneumonia}, \textit{Neumotórax} $\rightarrow$ \textbf{Pneumothorax}, \textit{Cardiomegalia} $\rightarrow$ \textbf{Cardiomegaly}), while excluding ambiguous terms lacking consistent mappings. For this dataset, we map \textit{Normal} to \textbf{No Finding}, and labels containing \textit{Mass} or \textit{Nodule} to \textbf{Lung Cancer}, again under radiologist guidance. Ambiguous or overly broad labels such as \textit{Infiltration} (ChestX-ray14) or \textit{Opacidad} (PadChest) were excluded to maintain semantic clarity. In cases where both \textbf{No Finding} and a pathology label co-occurred, the pathology label was prioritised. For studies with multiple co-occurring pathologies, all were retained as positive labels to reflect the multi-label nature of radiographic interpretation.

\begin{table}[h!]
\vspace{-5pt}
\begin{threeparttable}
\caption{\textbf{Macro-averaged AUROC and AUPRC scores} for all Train $\rightarrow$ Test settings. Datasets: MIMIC-CXR (MIM), CheXpert (CheX), PadChest (Pad), ChestX-ray14 (CXR). Models: ResNet50 (Res50), ResNet101 (Res101), ResNeXt101 (RX101), DenseNet161 (Dense), ConvNeXt-S (ConvS), ConvNeXt-B (ConvB), EfficientNetV2-S (Effic).}
\label{tab:macro_auroc_auprc}
\small
\begin{tabular}{lllllllll}
\hline
\textbf{Train $\rightarrow$ Test} & & \textbf{Res50} & \textbf{Res101} & \textbf{RX101} & \textbf{Dense} & \textbf{ConvS} & \textbf{ConvB} & \textbf{Effic} \\
\hline
CheX $\rightarrow$ CheX & \msff{\textbf{AUROC}}{\textbf{AUPRC}} & \msff{0.828}{0.524} & \msff{0.816}{0.508} & \msff{0.831}{0.541} & \msff{\textbf{0.848}}{\textbf{0.585}} & \msff{0.742}{0.321} & \msff{0.751}{0.346} & \msff{0.824}{0.544} \\
\cline{3-9}
CheX $\rightarrow$ MIM & \msff{\textbf{AUROC}}{\textbf{AUPRC}}  &  \msff{0.563}{0.174} & \msff{0.631}{0.214} & \msff{0.620}{0.199} & \msff{0.561}{0.174} & \msff{\textbf{0.696}}{\textbf{0.358}} & \msff{0.683}{0.303} & \msff{0.633}{0.207} \\
\cline{3-9}
CheX $\rightarrow$ CXR & \msff{\textbf{AUROC}}{\textbf{AUPRC}}  &  \msff{0.523}{0.175} & \msff{\textbf{0.613}}{\textbf{0.202}} & \msff{0.475}{0.162} & \msff{0.550}{0.185} & \msff{0.588}{0.182} & \msff{0.591}{0.191} & \msff{0.491}{0.155} \\
\cline{3-9}
CheX $\rightarrow$ Pad & \msff{\textbf{AUROC}}{\textbf{AUPRC}}  &  \msff{0.590}{0.191} & \msff{\textbf{0.880}}{\textbf{0.711}} & \msff{0.543}{0.173} & \msff{0.766}{0.405} & \msff{0.612}{0.187} & \msff{0.611}{0.203} & \msff{0.869}{0.679} \\
\hline
MIM $\rightarrow$ MIM  & \msff{\textbf{AUROC}}{\textbf{AUPRC}}  &  \msff{0.873}{0.667} & \msff{0.872}{0.663} & \msff{0.875}{0.669} & \msff{0.879}{0.670} & \msff{0.853}{0.629} & \msff{0.760}{0.362} & \msff{\textbf{0.893}}{\textbf{0.699}} \\ 
\cline{3-9}
MIM $\rightarrow$ CheX & \msff{\textbf{AUROC}}{\textbf{AUPRC}}  & \msff{0.516}{0.167} & \msff{0.498}{0.169} & \msff{0.508}{0.164} & \msff{0.501}{0.163} & \msff{0.551}{0.191} & \msff{\textbf{0.582}}{\textbf{0.266}} & \msff{0.510}{0.162} \\
\cline{3-9}
MIM $\rightarrow$ CXR & \msff{\textbf{AUROC}}{\textbf{AUPRC}}  &  \msff{0.505}{0.155} & \msff{0.499}{0.158} & \msff{0.487}{0.154} & \msff{0.478}{0.157} & \msff{0.708}{0.269} & \msff{\textbf{0.741}}{\textbf{0.301}} & \msff{0.475}{0.151} \\
\cline{3-9}
MIM $\rightarrow$ Pad & \msff{\textbf{AUROC}}{\textbf{AUPRC}}  & \msff{0.841}{0.600} & \msff{\textbf{0.860}}{\textbf{0.655}} & \msff{0.485}{0.170} & \msff{0.490}{0.172} & \msff{0.556}{0.165} & \msff{0.700}{0.316} & \msff{0.725}{0.252} \\
\hline
CXR $\rightarrow$ CXR & \msff{\textbf{AUROC}}{\textbf{AUPRC}}  &  \msff{0.914}{0.754} & \msff{0.913}{0.755} & \msff{0.913}{0.748} & \msff{\textbf{0.921}}{\textbf{0.770}} & \msff{0.881}{0.645} & \msff{0.869}{0.582} & \msff{0.920}{0.765} \\
\cline{3-9}
CXR $\rightarrow$ CheX & \msff{\textbf{AUROC}}{\textbf{AUPRC}}  & \msff{\textbf{0.538}}{0.184} & \msff{0.510}{0.163} & \msff{0.501}{0.173} & \msff{0.516}{0.182} & \msff{0.484}{0.166} & \msff{0.517}{0.180} & \msff{0.519}{\textbf{0.186}} \\
\cline{3-9}
CXR $\rightarrow$ MIM & \msff{\textbf{AUROC}}{\textbf{AUPRC}}  &  \msff{0.664}{0.223} & \msff{0.506}{0.167} & \msff{0.534}{0.176} & \msff{0.652}{0.218} & \msff{0.637}{0.210} & \msff{0.636}{0.230} & \msff{\textbf{0.685}}{\textbf{0.238}} \\
\cline{3-9}
CXR $\rightarrow$ Pad & \msff{\textbf{AUROC}}{\textbf{AUPRC}}  &  \msff{0.882}{0.725} & \msff{\textbf{0.895}}{\textbf{0.738}} & \msff{0.745}{0.267} & \msff{0.723}{0.247} & \msff{0.747}{0.363} & \msff{0.873}{0.693} & \msff{0.884}{0.701} \\
\hline
Pad $\rightarrow$ Pad & \msff{\textbf{AUROC}}{\textbf{AUPRC}}  & \msff{0.941}{0.811} & \msff{0.943}{0.817} & \msff{0.943}{0.817} & \msff{\textbf{0.944}}{\textbf{0.822}} & \msff{0.923}{0.716} & \msff{0.924}{0.727} & \msff{0.941}{0.815} \\
\cline{3-9}
Pad $\rightarrow$ CheX & \msff{\textbf{AUROC}}{\textbf{AUPRC}}  &  \msff{0.642}{\textbf{0.214}} & \msff{0.601}{0.192} & \msff{0.593}{0.195} & \msff{\textbf{0.646}}{0.206} & \msff{0.533}{0.182} & \msff{0.533}{0.180} & \msff{0.566}{0.183} \\
\cline{3-9}
Pad $\rightarrow$ MIM & \msff{\textbf{AUROC}}{\textbf{AUPRC}}  & \msff{\textbf{0.681}}{\textbf{0.249}} & \msff{0.562}{0.189} & \msff{0.538}{0.184} & \msff{0.578}{0.188} & \msff{0.514}{0.163} & \msff{0.493}{0.164} & \msff{0.480}{0.163} \\
\cline{3-9}
Pad $\rightarrow$ CXR & \msff{\textbf{AUROC}}{\textbf{AUPRC}}  & \msff{\textbf{0.830}}{0.561} & \msff{0.817}{0.558} & \msff{0.806}{0.508} & \msff{0.814}{0.559} & \msff{0.808}{0.566} & \msff{0.808}{\textbf{0.570}} & \msff{0.807}{0.529} \\
\hline
\end{tabular}
\end{threeparttable}
\vspace{-5pt}
\end{table}

\newcommand{\heatmap}[2]{%
  \begin{subfigure}{0.33\textwidth}\centering
    \includegraphics[width=\linewidth]{#1}
    \subcaption{#2}
  \end{subfigure}%
}

\begin{figure}[h!]
\vspace{-10pt}
  \centering
  \captionsetup[sub]{labelformat=empty,justification=centering}

  \heatmap{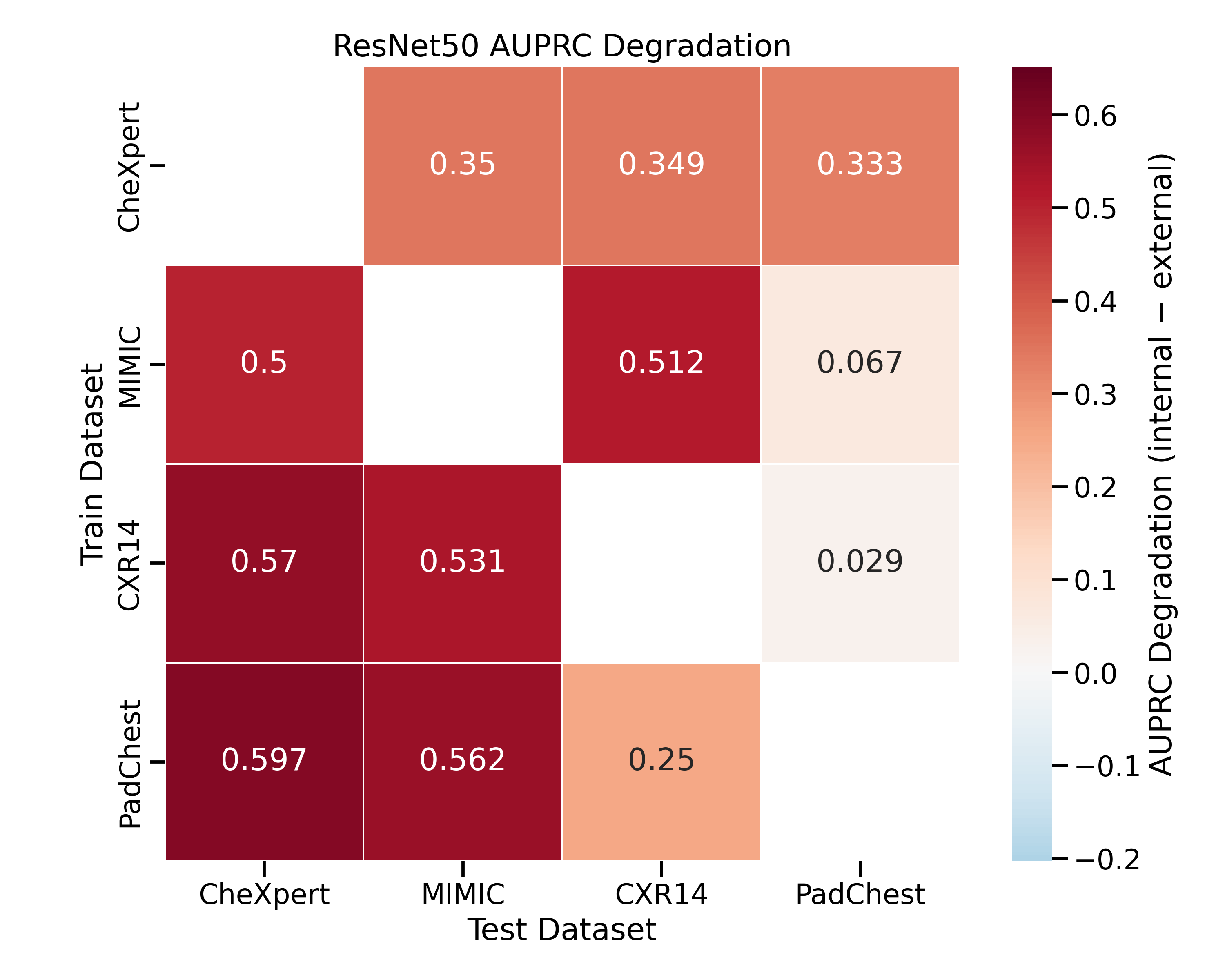}{ResNet50}\hfill
  \heatmap{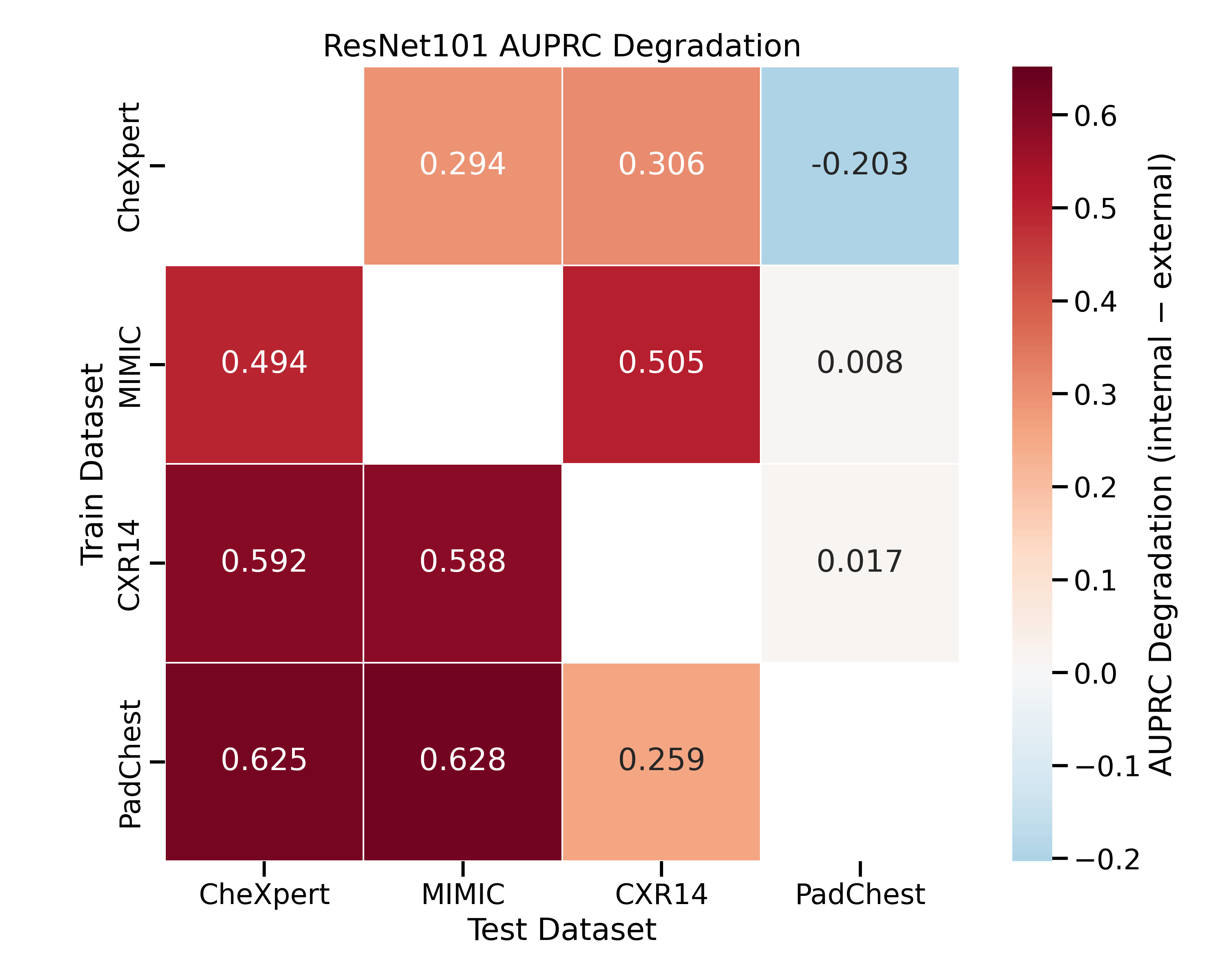}{ResNet101}\hfill
  \heatmap{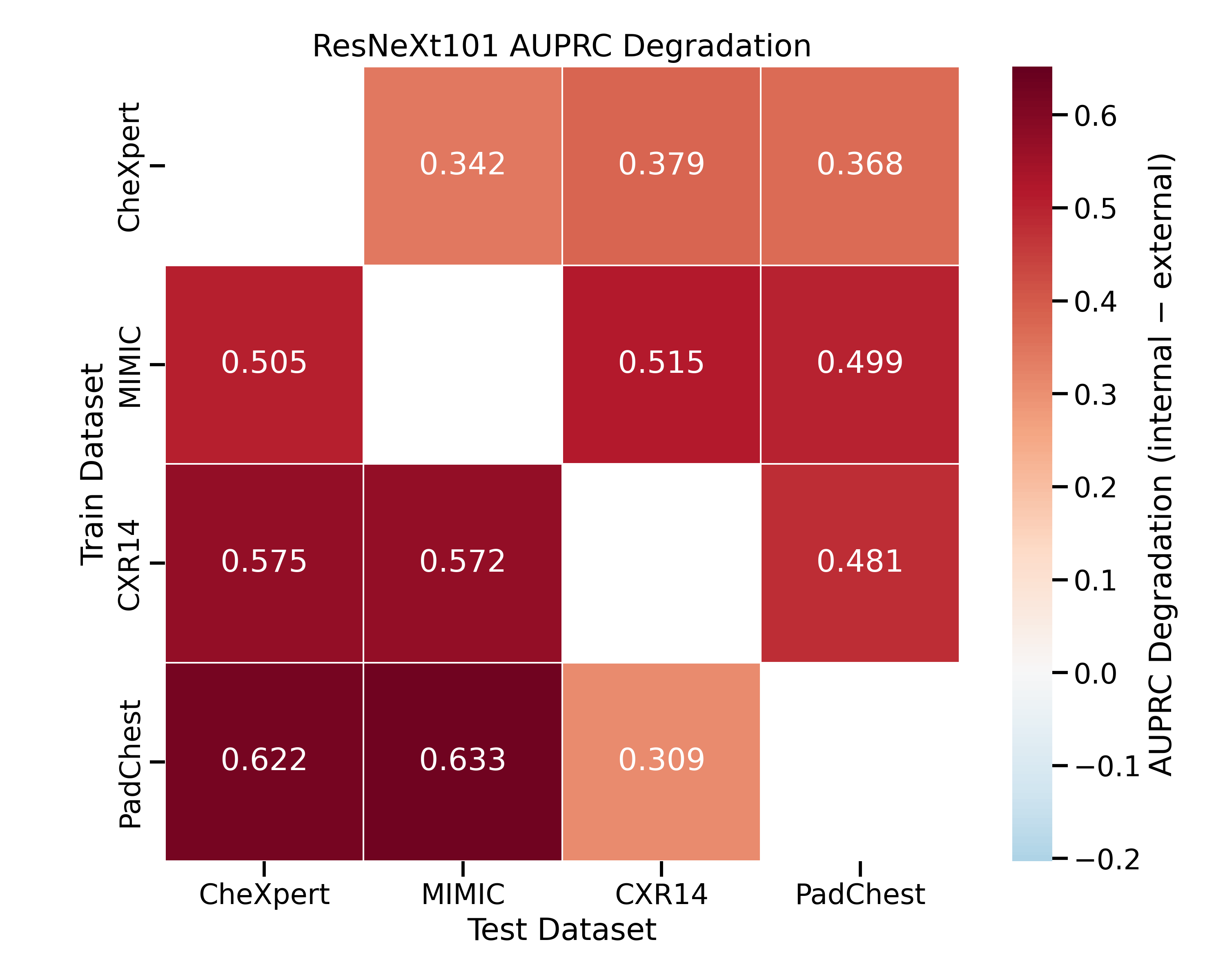}{ResNeXt101}\hfill
  \heatmap{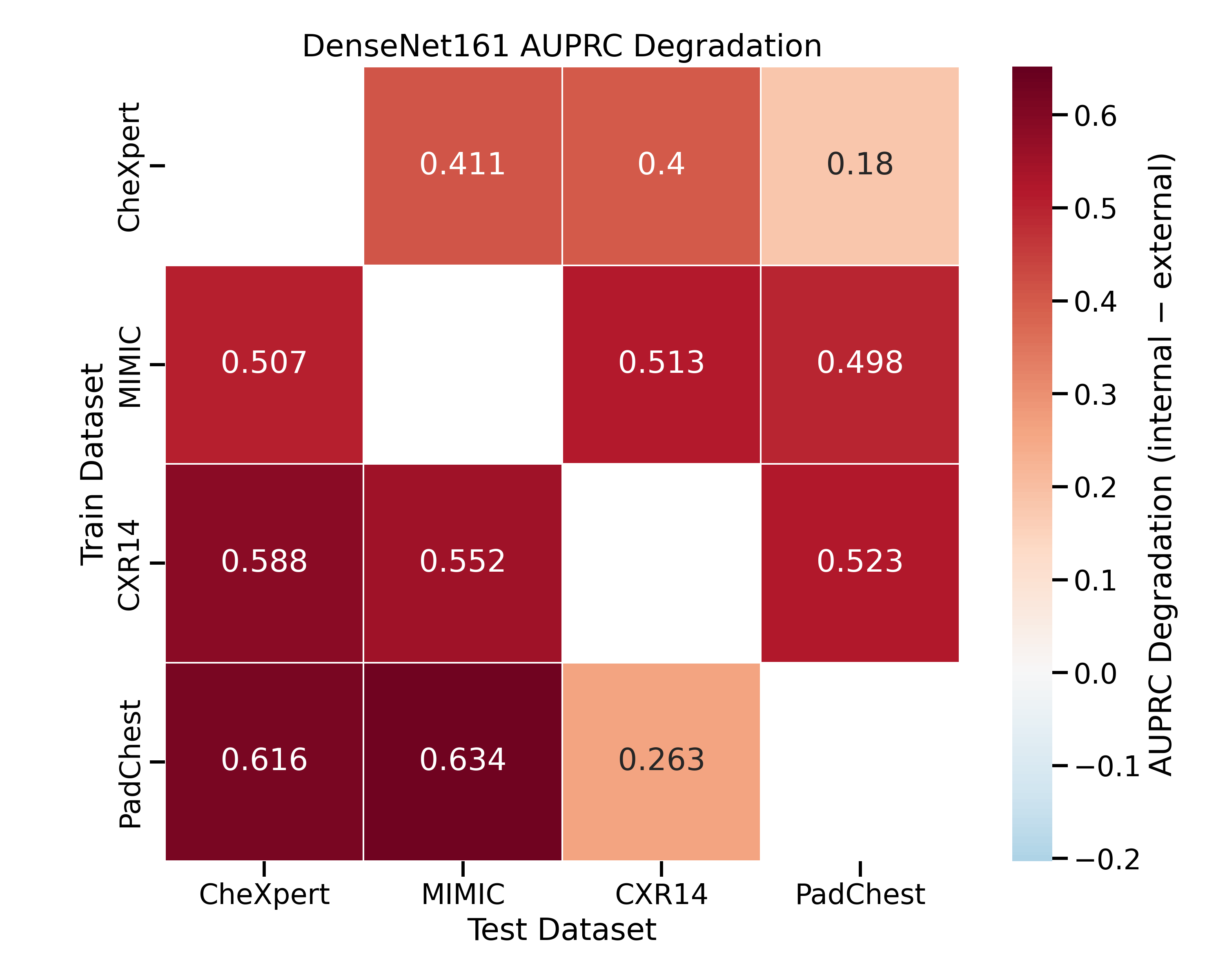}{DenseNet161}\hfill
  \heatmap{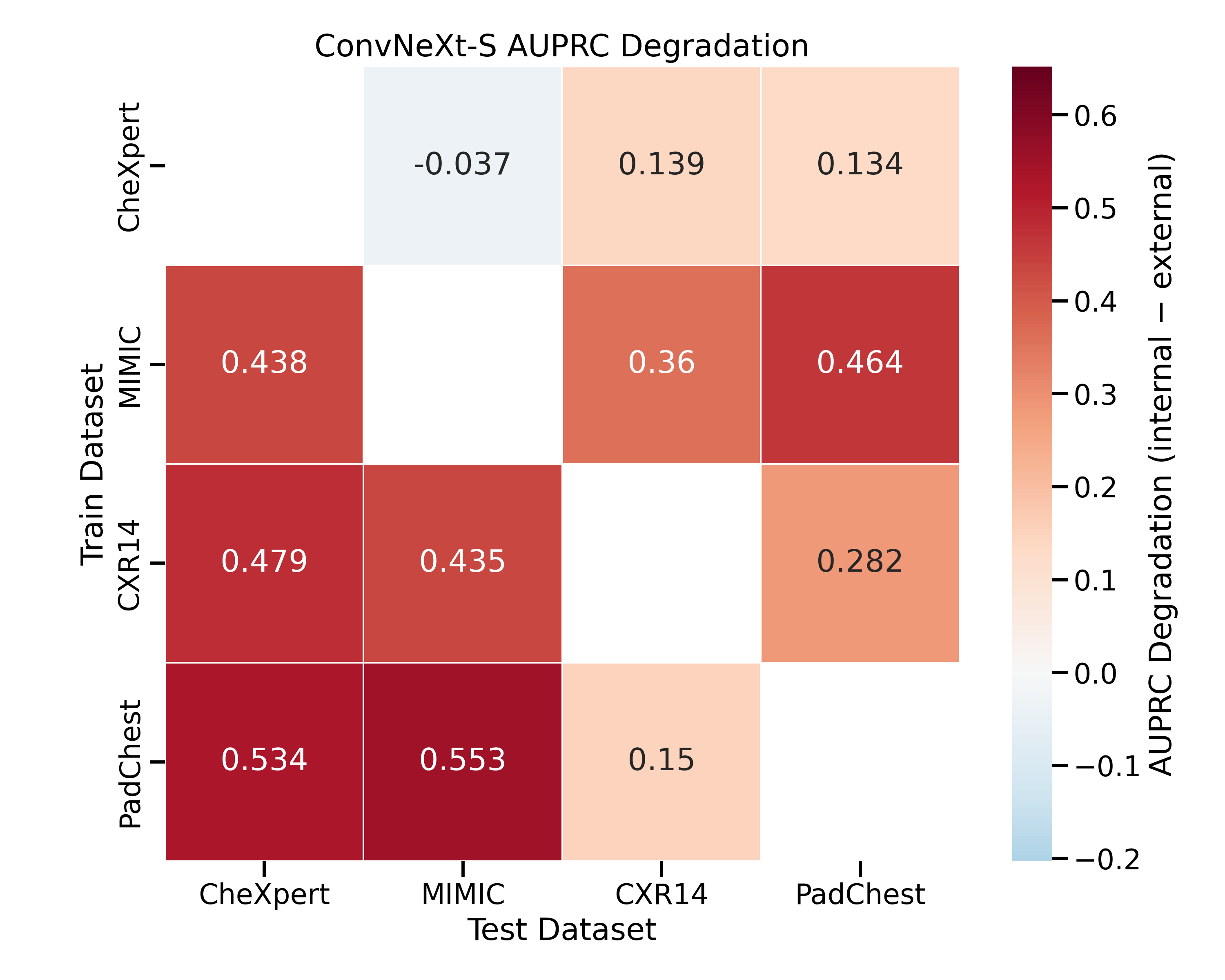}{ConvNeXt-S}\hfill
  \heatmap{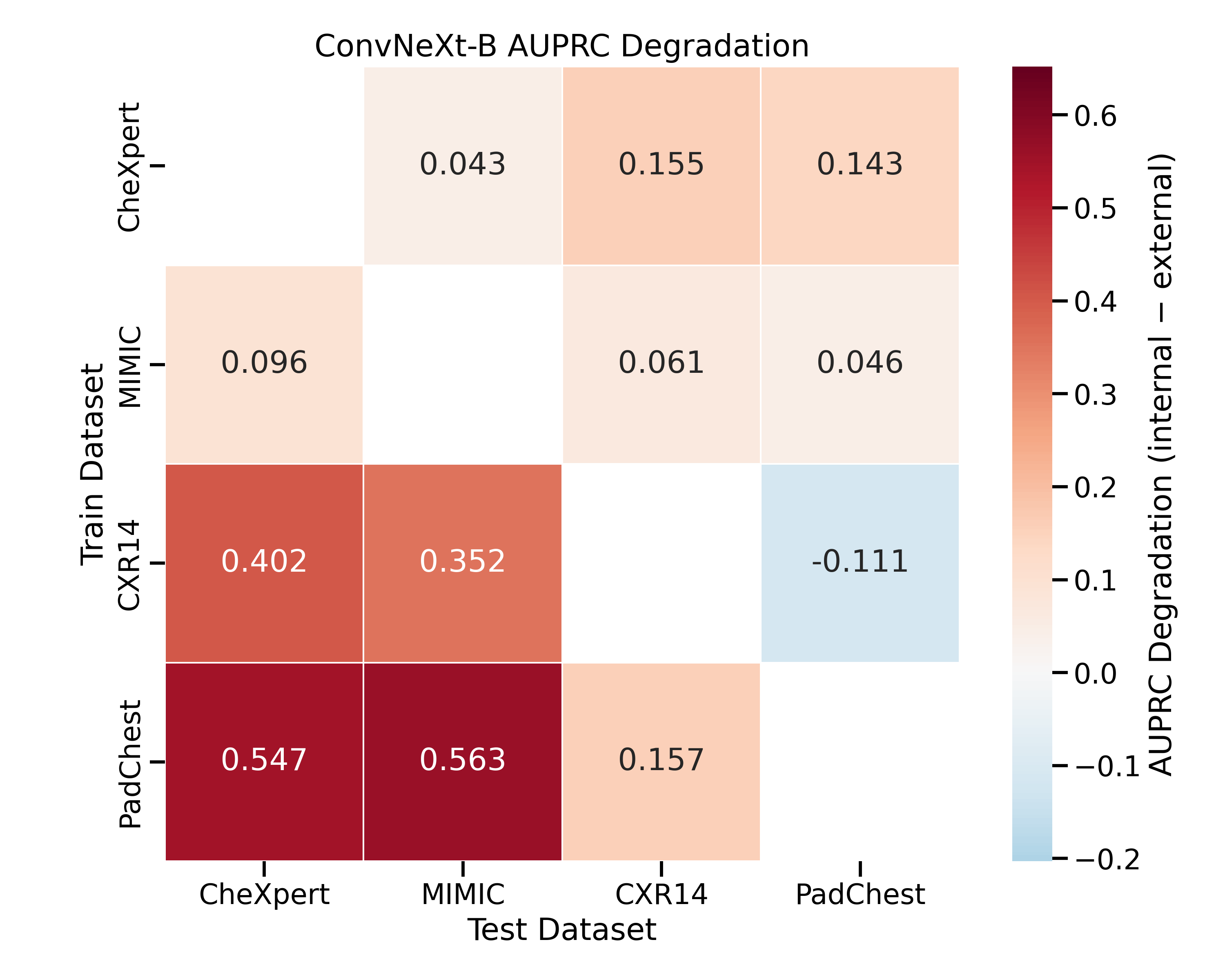}{ConvNeXt-B}\hfill
  \heatmap{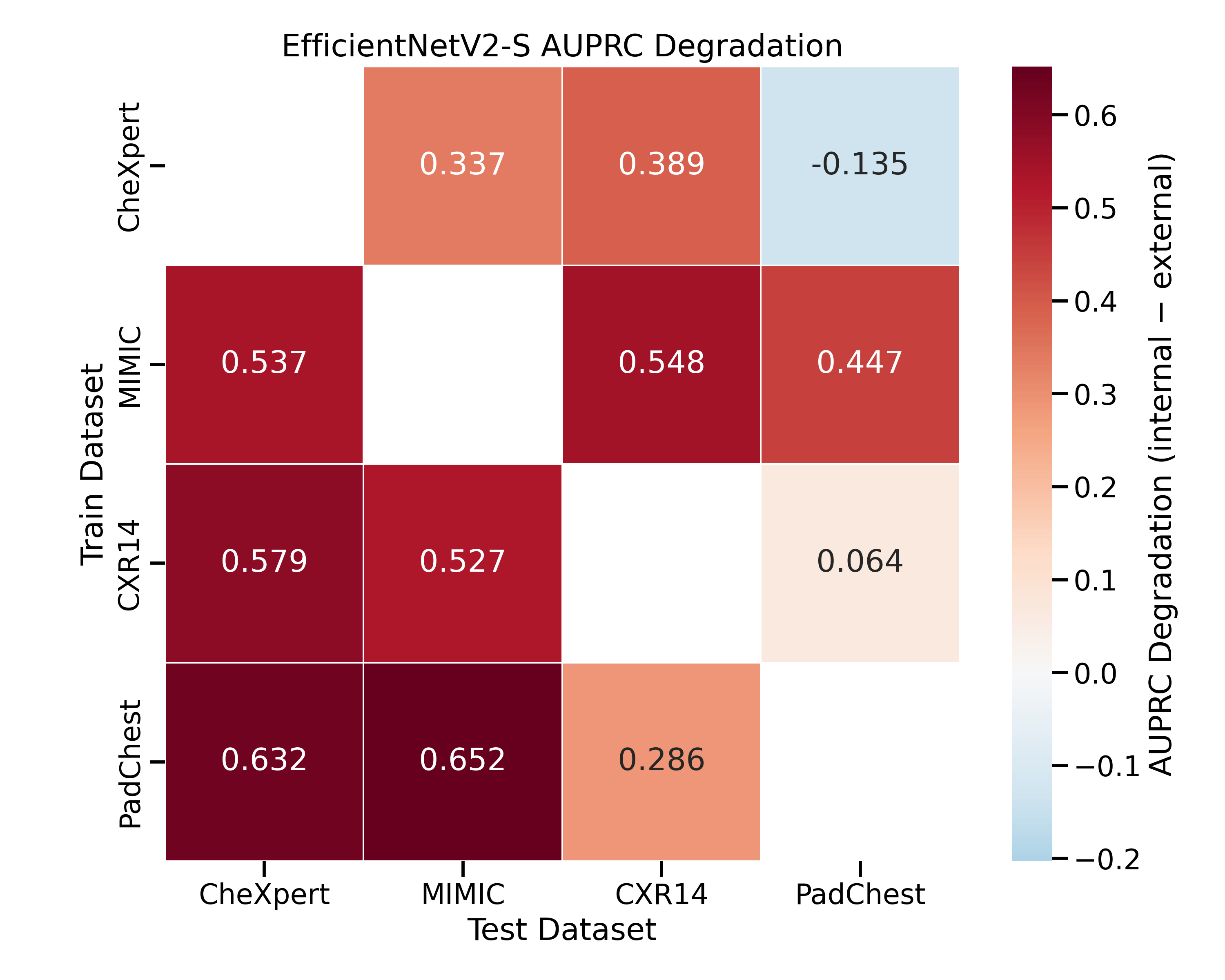}{EfficientNetV2-S}\hfill

  \caption{Macro-averaged \textbf{AUPRC degradation} heatmaps for each model architecture (internal $-$ external). Rows = training dataset; columns = test dataset. Shared color scale across panels.}
  \label{fig:auprc_heatmaps}
  \vspace{-10pt}
\end{figure}

\vspace{-5pt}
\subsection{Evaluation Metrics}

Model performance is assessed using both threshold-independent and threshold-dependent metrics to capture complementary aspects of diagnostic performance. 
For threshold-independent evaluation, we compute the area under the receiver operating characteristic curve (AUROC) and the area under the precision–recall curve (AUPRC) of each model architecture. AUROC reflects a model’s ability to distinguish between positive and negative cases across the full range of thresholds, providing an overall measure of classification performance. However, it can be misleading in medical imaging tasks with highly imbalanced labels \cite{auroc}. In contrast, AUPRC is more informative under class imbalance, as it emphasises performance on the positive class and better reflects model behaviour on rarer conditions such as pneumothorax. 
Results are reported using macro-averaged AUROC, AUPRC, and threshold-dependent metrics to ensure that domain shift effects are assessed uniformly across all pathologies. This avoids conclusions being dominated by a small number of high-prevalence findings and enables consistent comparison across datasets with differing label distributions. Given our multi-architecture evaluation, number of pathologies, and large number of train--test combinations, macro-averaging provides a clear and interpretable summary of systematic generalisation behaviour without overcomplicating results. This choice is also aligned with our label harmonisation procedure (Section~\ref{sec:labelharm}) and the substantial class-imbalance differences across datasets (Table~\ref{tab:classdist}), ensuring that comparisons reflect consistent pathology definitions and are not dominated by majority labels or dataset-specific prevalences.
We show macro-averaged AUROC and AUPRC scores for each model architecture, for each training-testing dataset pair, in Table \ref{tab:macro_auroc_auprc}. AUPRC degradation between internal and external datasets is summarised in Figure \ref{fig:auprc_heatmaps}. 

\begin{figure}[h!]
    \centering
    \vspace{-10pt}
    \includegraphics[width=0.7\linewidth]{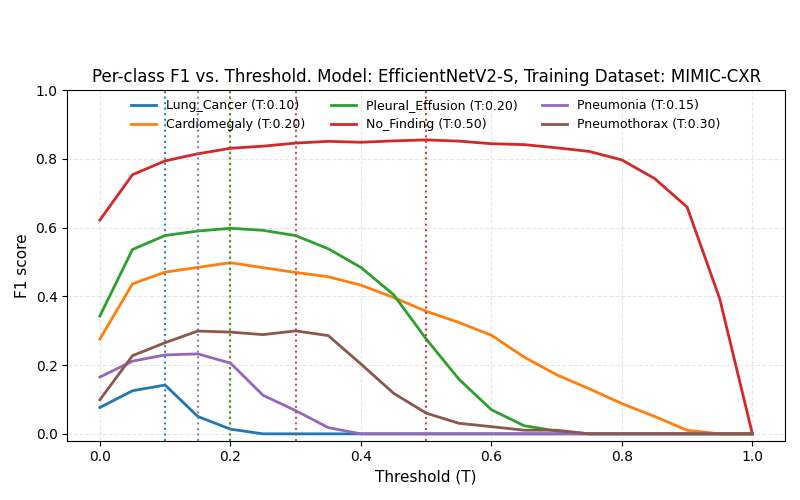}
    \caption{Example by-class F1 threshold selection, for the EfficientNetV2-S model trained on the MIMIC-CXR dataset. This is repeated for all architectures, on each training dataset. These internal thresholds are used to assess model performance on external test sets.}
    \label{fig:f1_curve}
    \vspace{-10pt}
\end{figure}

\begin{figure}[h!]
\vspace{-8pt}
  \centering
  \captionsetup[sub]{labelformat=empty,justification=centering}

  \heatmap{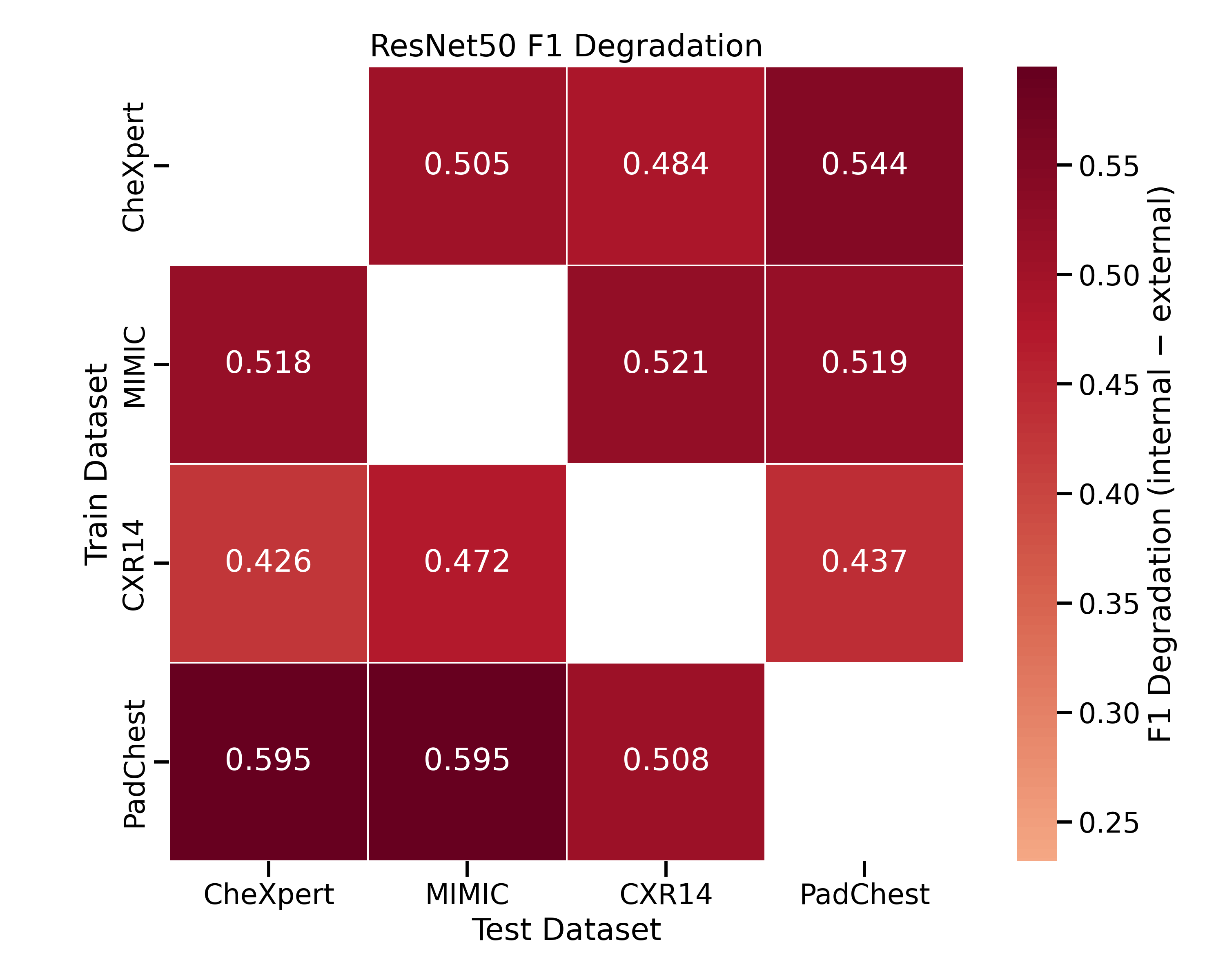}{ResNet50}\hfill
  \heatmap{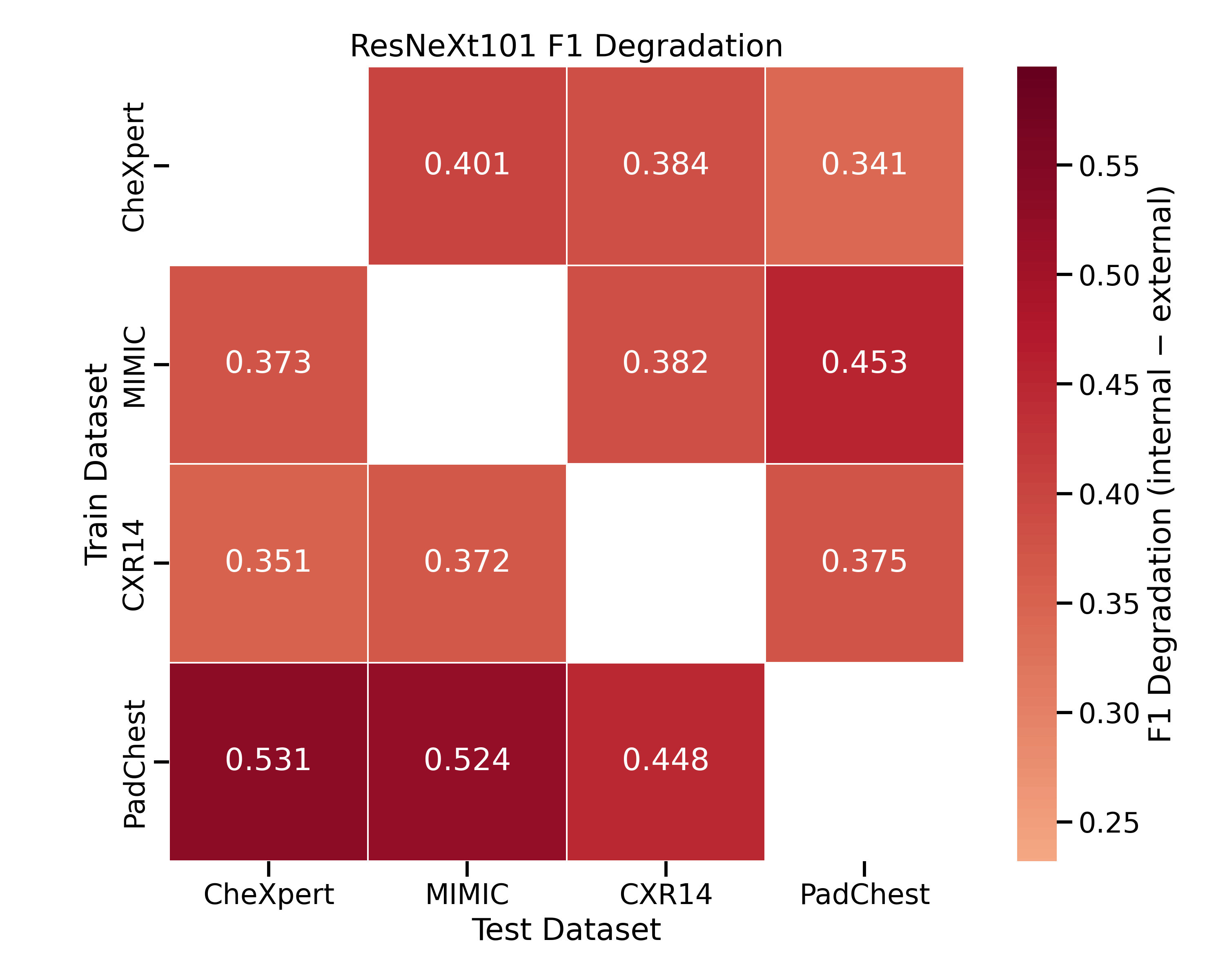}{ResNet101}\hfill
  \heatmap{ResNeXt101_f1_degradation_heatmap.png}{ResNeXt101}\hfill
  \heatmap{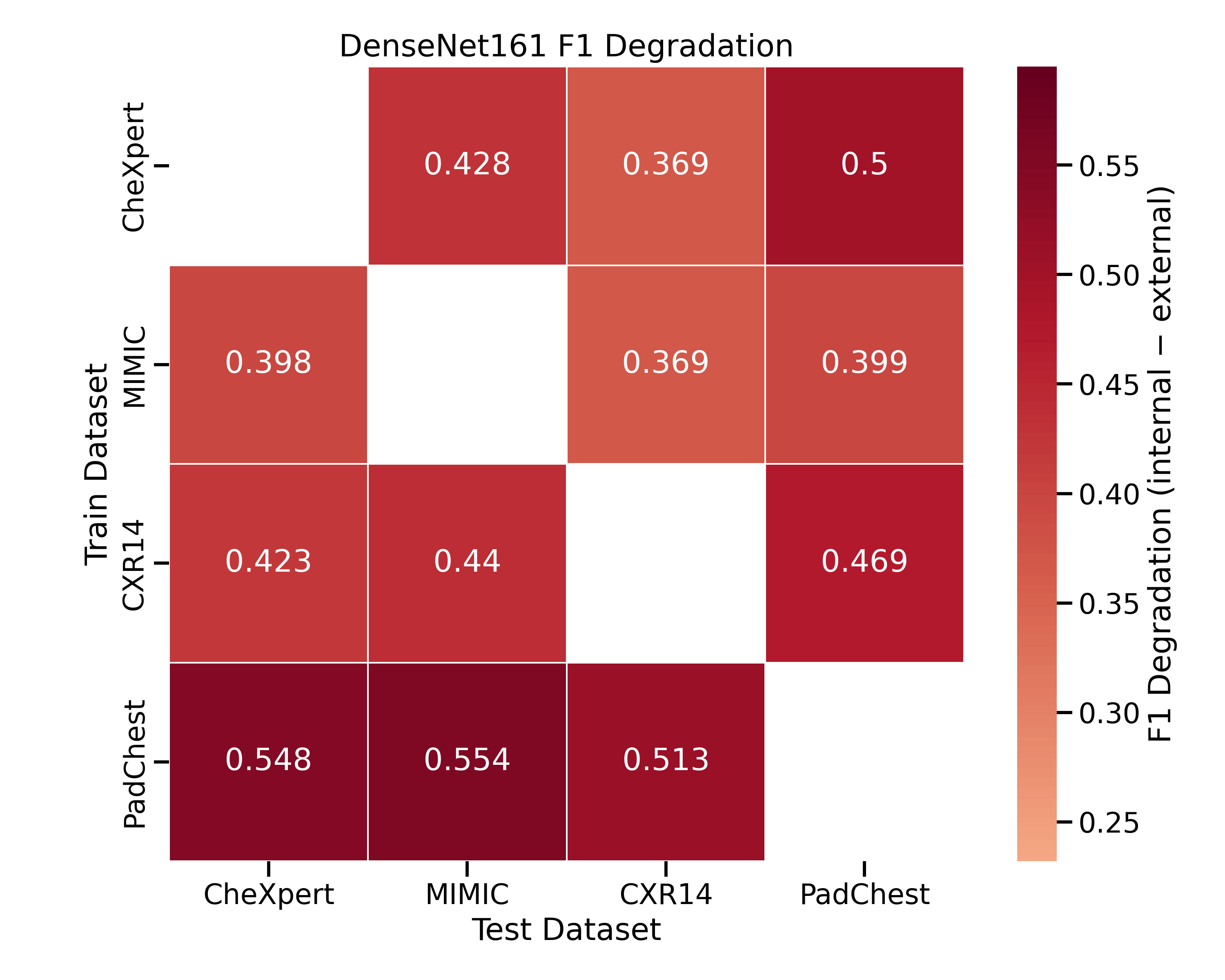}{DenseNet161}\hfill
  \heatmap{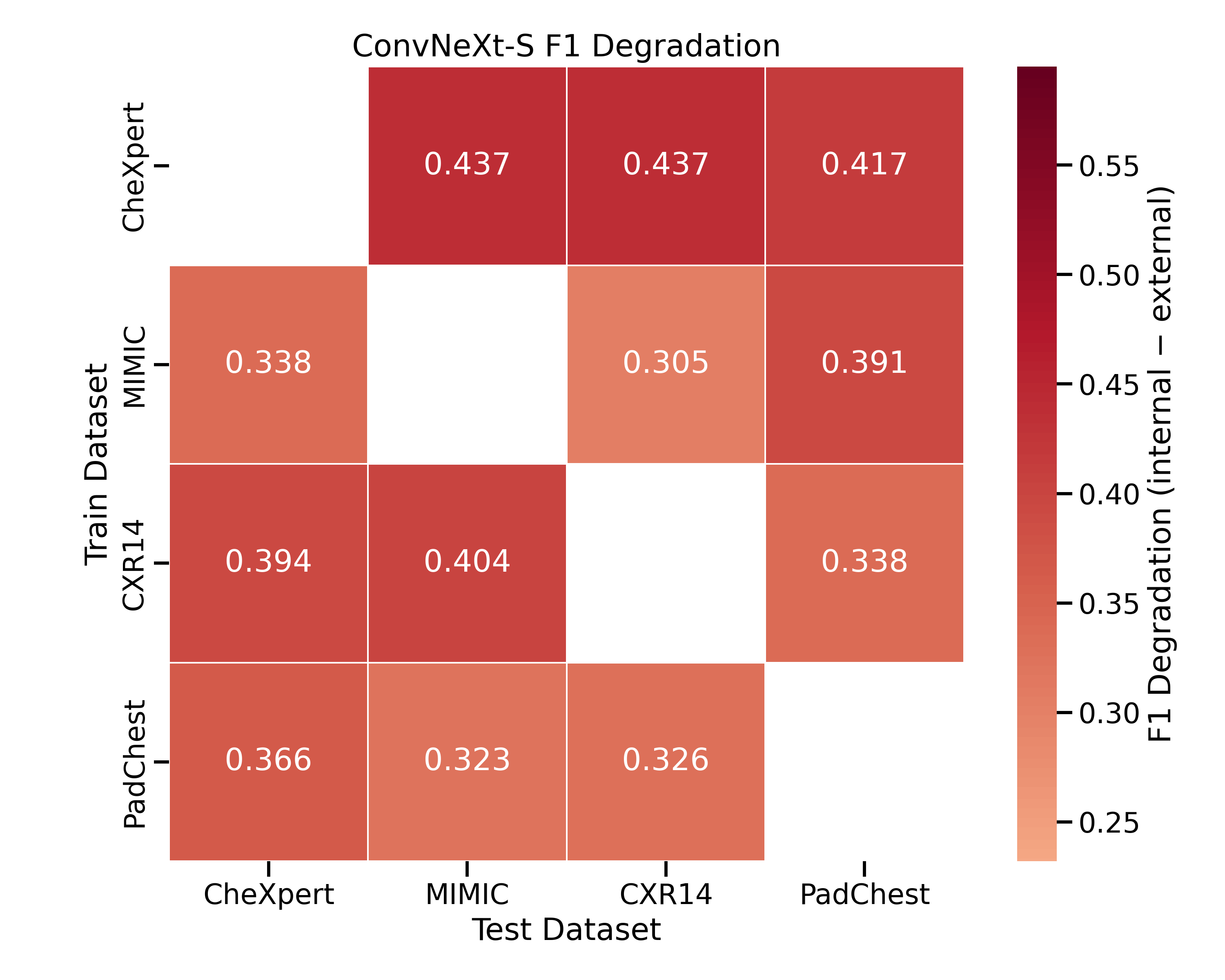}{ConvNeXt-S}\hfill
  \heatmap{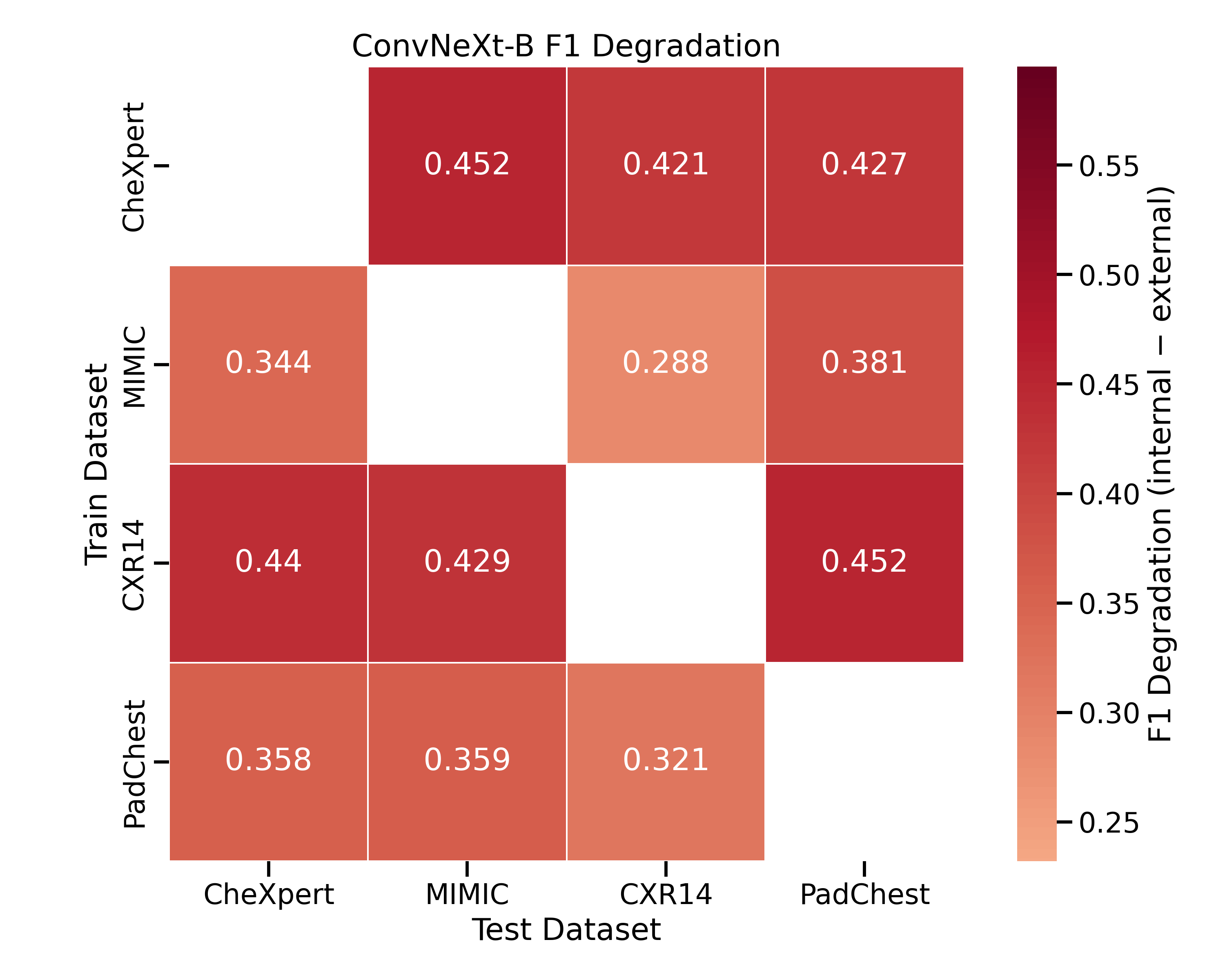}{ConvNeXt-B}\hfill
  \heatmap{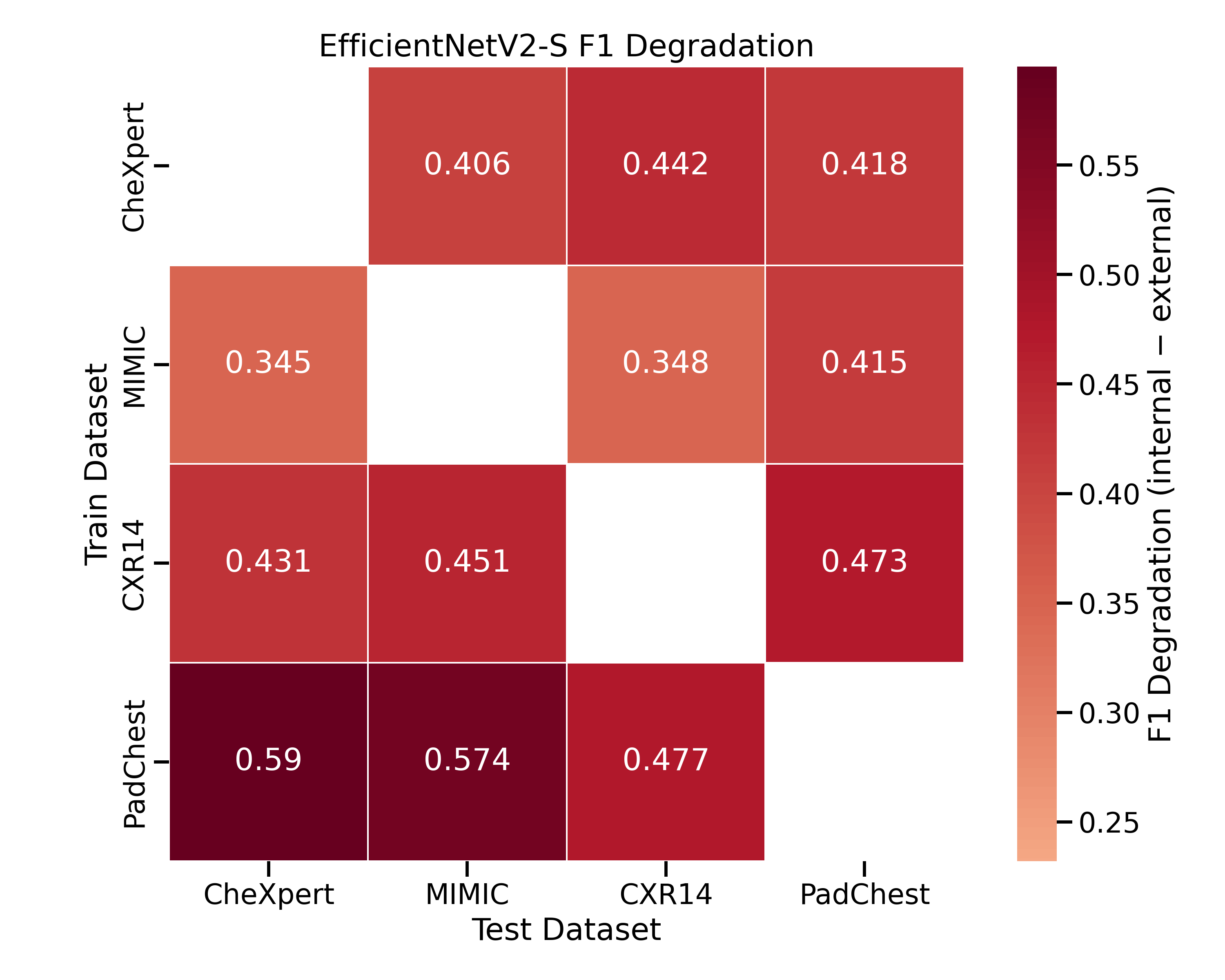}{EfficientNetV2-S}\hfill

  \caption{Macro-averaged \textbf{F1 score degradation} heatmaps (internal $-$ external) for all model architectures. Rows = training dataset; columns = test dataset. Shared color scale across panels.}
  \label{fig:f1_degradation_heatmaps}
  \vspace{-8pt}
\end{figure}

For threshold-dependent evaluation, we report Sensitivity (true positive rate), Specificity (true negative rate), and F1 score. For each model, class-specific decision thresholds are tuned on the training dataset’s validation split to maximise per-class F1, then fixed and applied unchanged to all external test datasets. An example threshold selection process is shown in Figure \ref{fig:f1_curve} for the EfficientNetV2-S model, trained on MIMIC-CXR. We summarise model performance as macro-averaged F1, Sensitivity, and Specificity computed with these fixed thresholds, with results shown in Table \ref{tab:macro_sens_spec_f1}. By selecting thresholds on the internal (in-distribution) data and evaluating them without recalibration externally, we specifically quantify performance degradation caused by domain shift—the generalisation gap between internal and external datasets. We summarise the F1 score degradation of models between internal and external datasets in Figure \ref{fig:f1_degradation_heatmaps}. This evaluation reflects realistic deployment scenarios, where recalibration of decision thresholds is rarely possible when applying models across institutions. 
Reporting both general (AUROC, AUPRC) and threshold-dependent metrics (Sensitivity, Specificity, F1 score) provides a thorough and clinically meaningful assessment of external model generalisability and domain shift.

\vspace{-5pt}
\subsection{Results}

Across all architectures and datasets, internal (in-distribution) performance is consistently strong, while external (out-of-distribution) performance drops significantly. The reduction is larger and more frequent for AUPRC and F1 than for AUROC, indicating that domain shift impacts calibration and positive-class retrieval more than rank ordering.

\vspace{-5pt}
\subsubsection{Threshold-independent performance (AUROC / AUPRC)}
Within-dataset (internal) performance is uniformly high across all model architectures. DenseNet161 is the highest performing model in the CheXpert, ChestX-ray14 and PadChest internal cases, with AUROC/AUPRC scores of 0.848/0.585, 0.921/0.770 and 0.944/0.822 respectively. EfficientNetV2-S is the highest performing model in the MIMIC-CXR internal case, with an AUROC/AUPRC score of 0.893/0.699. 
Cross-dataset (external) evaluations show systematic degradation, most notably in AUPRC scores. For example, models trained on the CheXpert dataset see AUPRC degradations of up to 0.411 when evaluated on external datasets. The most notable degradations occur on models trained on PadChest, which are then applied to CheXpert or MIMIC-CXR – AUPRC scores decrease by more than 0.5 for all model architectures. AUROC degradation in these cases is also significant, in the order of 0.2 – 0.3.
While no model architecture dominates in all cases, we can extract some trends from Table \ref{tab:macro_auroc_auprc} and Figure \ref{fig:auprc_heatmaps}. DenseNet161 and EfficientNetV2-S tend to be strongest internally. In external cases, ConvNeXt-B and ConvNeXt-S perform notably well when trained on MIMIC-CXR, while ResNet50 and ResNet101 perform well when trained on CheXpert, ChestX-ray14 or PadChest. We also note that models trained on ChestX-ray14 or CheXpert transfer best to PadChest, while external models evaluated on CheXpert often exhibit the most significant drops in AUPRC scores.

\newcommand{\msf}[3]{\makecell{ #1\\ #2\\ #3}}

\begin{table}[h!]
\vspace{-5pt}
\begin{threeparttable}
\caption{\small\textbf{Sensitivity (Sens), Specificity (Spec), and F1} for all Train $\rightarrow$ Test settings. Thresholds chosen on training dataset’s validation split and applied unchanged to test dataset. Datasets: MIMIC-CXR (MIM), CheXpert (CheX), PadChest (Pad), ChestX-ray14 (CXR). Models: ResNet50 (Res50), ResNet101 (Res101), ResNeXt101 (RX101), DenseNet161 (Dense), ConvNeXt-S (ConvS), ConvNeXt-B (ConvB), EfficientNetV2-S (Effic).}
\label{tab:macro_sens_spec_f1}
\small
\begin{tabular}{lllllllll}
\hline
\textbf{Train $\rightarrow$ Test} &&  \textbf{Res50} & \textbf{Res101} & \textbf{RX101} & \textbf{Dense} & \textbf{ConvS} & \textbf{ConvB} & \textbf{Effic} \\
\hline
CheX $\rightarrow$ CheX & \msf{\textbf{Sens}}{\textbf{Spec}}{\textbf{F1}} & \msf{0.838}{\textbf{0.786}}{\textbf{0.846}} & \msf{0.753}{0.564}{0.683} & \msf{0.841}{0.651}{0.725} & \msf{0.708}{0.717}{0.754} & \msf{\textbf{0.894}}{0.629}{0.704} & \msf{0.862}{0.645}{0.689} & \msf{0.827}{0.676}{0.735} \\
\cline{3-9}
CheX $\rightarrow$ MIM & \msf{\textbf{Sens}}{\textbf{Spec}}{\textbf{F1}} & \msf{0.899}{\textbf{0.323}}{\textbf{0.341}} & \msf{0.896}{0.259}{0.320} & \msf{0.922}{0.246}{0.324} & \msf{\textbf{0.931}}{0.242}{0.326} & \msf{0.904}{0.214}{0.267} & \msf{0.922}{0.208}{0.237} & \msf{0.894}{0.293}{0.329} \\
\cline{3-9}
CheX $\rightarrow$ CXR& \msf{\textbf{Sens}}{\textbf{Spec}}{\textbf{F1}} & \msf{0.854}{0.426}{0.362} & \msf{0.968}{0.312}{0.358} & \msf{0.900}{0.323}{0.341} & \msf{0.907}{\textbf{0.440}}{\textbf{0.385}} & \msf{0.877}{0.102}{0.267} & \msf{0.891}{0.173}{0.268} & \msf{\textbf{0.992}}{0.044}{0.293} \\
\cline{3-9}
CheX $\rightarrow$ Pad& \msf{\textbf{Sens}}{\textbf{Spec}}{\textbf{F1}} & \msf{0.980}{0.097}{0.302} & \msf{0.989}{0.100}{0.305} & \msf{0.978}{\textbf{0.376}}{\textbf{0.384}} & \msf{0.982}{0.091}{0.254} & \msf{\textbf{0.992}}{0.003}{0.287} & \msf{0.936}{0.009}{0.262} & \msf{0.983}{0.156}{0.317} \\
\hline
MIM $\rightarrow$ MIM& \msf{\textbf{Sens}}{\textbf{Spec}}{\textbf{F1}} & \msf{\textbf{0.886}}{\textbf{0.883}}{\textbf{0.867}} & \msf{0.739}{0.745}{0.790} & \msf{0.757}{0.769}{0.720} & \msf{0.744}{0.784}{0.727} & \msf{0.795}{0.549}{0.693} & \msf{0.712}{0.593}{0.645} & \msf{0.742}{0.864}{0.712} \\ 
\cline{3-9}
MIM $\rightarrow$ CheX& \msf{\textbf{Sens}}{\textbf{Spec}}{\textbf{F1}} & \msf{0.914}{0.335}{0.349} & \msf{0.915}{0.351}{0.355} & \msf{0.929}{0.316}{0.347} & \msf{\textbf{0.936}}{0.250}{0.329} & \msf{0.879}{0.386}{0.355} & \msf{0.894}{0.313}{0.301} & \msf{0.909}{\textbf{0.393}}{\textbf{0.367}} \\
\cline{3-9}
MIM $\rightarrow$ CXR& \msf{\textbf{Sens}}{\textbf{Spec}}{\textbf{F1}} & \msf{\textbf{0.979}}{0.263}{0.346} & \msf{0.970}{0.319}{0.361} & \msf{0.975}{0.241}{0.338} & \msf{0.950}{0.327}{0.358} & \msf{0.946}{\textbf{0.414}}{\textbf{0.388}} & \msf{0.965}{0.312}{0.357} & \msf{0.947}{0.350}{0.364} \\
\cline{3-9}
MIM $\rightarrow$ Pad& \msf{\textbf{Sens}}{\textbf{Spec}}{\textbf{F1}} & \msf{0.899}{0.247}{\textbf{0.348}} & \msf{0.931}{\textbf{0.277}}{0.314} & \msf{0.999}{0.008}{0.267} & \msf{0.960}{0.221}{0.328} & \msf{0.985}{0.095}{0.302} & \msf{\textbf{1.000}}{0.003}{0.264} & \msf{0.995}{0.088}{0.297} \\
\hline
CXR $\rightarrow$ CXR& \msf{\textbf{Sens}}{\textbf{Spec}}{\textbf{F1}} &\msf{0.808}{0.755}{0.758} & \msf{0.787}{0.842}{0.711} & \msf{0.803}{0.659}{0.658} & \msf{0.787}{\textbf{0.878}}{0.756} & \msf{0.846}{0.556}{0.634} & \msf{\textbf{0.892}}{0.641}{0.698} & \msf{0.870}{0.769}{\textbf{0.763}} \\
\cline{3-9}
CXR $\rightarrow$ CheX& \msf{\textbf{Sens}}{\textbf{Spec}}{\textbf{F1}} &\msf{0.921}{0.275}{0.332} & \msf{0.922}{0.319}{\textbf{0.346}} & \msf{0.958}{0.142}{0.307} & \msf{0.862}{\textbf{0.335}}{0.333} & \msf{0.896}{0.091}{0.240} & \msf{\textbf{1.000}}{0.003}{0.258} & \msf{0.934}{0.260}{0.332} \\
\cline{3-9}
CXR $\rightarrow$ MIM& \msf{\textbf{Sens}}{\textbf{Spec}}{\textbf{F1}} & \msf{\textbf{0.999}}{0.001}{0.286} & \msf{0.960}{0.069}{0.290} & \msf{\textbf{0.999}}{0.002}{0.286} & \msf{0.957}{\textbf{0.178}}{\textbf{0.316}} & \msf{0.997}{0.006}{0.230} & \msf{0.998}{0.003}{0.269} & \msf{0.973}{0.145}{0.312} \\
\cline{3-9}
CXR $\rightarrow$ Pad& \msf{\textbf{Sens}}{\textbf{Spec}}{\textbf{F1}} &\msf{0.982}{\textbf{0.175}}{\textbf{0.321}} & \msf{0.991}{0.089}{0.268} & \msf{0.940}{0.078}{0.283} & \msf{0.999}{0.005}{0.287} & \msf{0.989}{0.061}{0.296} & \msf{\textbf{1.000}}{0.001}{0.246} & \msf{0.961}{0.004}{0.290} \\
\hline
Pad$\rightarrow$ Pad& \msf{\textbf{Sens}}{\textbf{Spec}}{\textbf{F1}} &\msf{0.923}{0.716}{\textbf{0.881}} & \msf{0.920}{0.703}{0.818} & \msf{0.930}{\textbf{0.794}}{0.817} & \msf{0.915}{0.715}{0.822} & \msf{\textbf{0.993}}{0.612}{0.612} & \msf{0.971}{0.587}{0.604} & \msf{0.907}{0.714}{0.875} \\
\cline{3-9}
Pad $\rightarrow$ CheX& \msf{\textbf{Sens}}{\textbf{Spec}}{\textbf{F1}} & \msf{0.997}{0.005}{0.286} & \msf{0.991}{0.032}{\textbf{0.290}} & \msf{0.999}{0.002}{0.286} & \msf{0.994}{\textbf{0.045}}{0.274} & \msf{\textbf{1.000}}{0.003}{0.246} & \msf{\textbf{1.000}}{0.004}{0.246} & \msf{0.992}{0.007}{0.285} \\
\cline{3-9}
Pad $\rightarrow$ MIM& \msf{\textbf{Sens}}{\textbf{Spec}}{\textbf{F1}} & \msf{0.994}{0.008}{0.286} & \msf{0.998}{0.003}{\textbf{0.586}} & \msf{0.991}{0.009}{0.293} & \msf{0.993}{0.052}{0.268} & \msf{0.990}{0.082}{0.289} & \msf{\textbf{1.000}}{0.001}{0.245} & \msf{0.982}{\textbf{0.094}}{0.301} \\
\cline{3-9}
Pad $\rightarrow$ CXR& \msf{\textbf{Sens}}{\textbf{Spec}}{\textbf{F1}} & \msf{0.964}{0.358}{0.373} & \msf{0.977}{0.222}{0.334} & \msf{0.967}{0.345}{0.369} & \msf{0.990}{0.118}{0.309} & \msf{0.982}{0.102}{0.286} & \msf{\textbf{0.993}}{0.003}{0.283} & \msf{0.914}{\textbf{0.371}}{\textbf{0.398}} \\
\hline
\end{tabular}
\end{threeparttable}
\vspace{-5pt}
\end{table}

\vspace{-5pt}
\subsubsection{Threshold-dependent performance (Sensitivity / Specificity / F1)}

With thresholds fixed per class on the internal validation split and applied unchanged externally, most external settings display very high Sensitivity but very low Specificity, indicating miscalibration under domain shift. This pattern yields substantially lower external F1 despite large true positive rates.
Internal performance is again relatively high across all model architectures. ResNet50 yields the highest F1 scores across internal MIMIC-CXR, CheXpert and PadChest cases (0.867, 0.846, 0.881 respectively), while EfficientNetV2-S has the highest performance for ChestX-ray14 (0.763).
In external cases, multiple models exhibit extremely high Sensitivity scores (>0.9) with Specificity close to 0, leading to low F1 scores in the range of 0.2-0.3. For example, ResNet50, which is the highest performing internal model for the ChestX-ray14 dataset, drops performance dramatically, with an F1 score of 0.286 (Sensitivity 0.001, Specificity 0.999) when evaluated on MIMIC-CXR. This trend is shown across multiple external cases across each dataset, and is especially common for the ConvNeXt-S and ConvNeXt-B models.

\vspace{-5pt}
\subsection{Summary of Findings}

Our cross-dataset evaluation highlights the significant impact of domain shift on chest radiography models. While all architectures achieved strong internal performance, external generalisation was consistently poor, with much sharper declines in AUPRC and F1 than in AUROC.  Threshold-dependent analysis revealed that models typically retained high sensitivity but lost specificity under domain shift, indicating systematic miscalibration across institutions. Importantly, no single architecture proved consistently robust: performance degradation depended more strongly on the source–target dataset pair than on model choice. These results emphasise that dataset-specific characteristics, rather than network design, are the primary drivers of cross-institutional generalisation failure in chest radiography.
In practical deployment scenarios where decision thresholds are fixed and recalibration is not feasible, these degradations suggest that models trained on a single institutional dataset may produce unreliable predictions when applied externally, particularly for rarer pathologies where precision--recall performance is critical. Such shifts may increase false positives or false negatives, potentially affecting diagnostic confidence and downstream clinical decision-making, underscoring the need for rigorous external validation prior to clinical use.

\vspace{-5pt}
\section{Empirical Analysis: Dataset Bias}
In addition to domain shift, deep learning models are also vulnerable to dataset bias, where differences in patient populations or acquisition metadata may reduce generalisability. We analyse the impact of these demographic characteristics across our public chest radiography datasets, specifically whether dataset-specific signatures are learnable directly from images, and whether there are significant subgroup disparities in model performance.

We first examine the available demographic information for each dataset. Age and sex are reported in CheXpert, PadChest, and ChestX-ray14, but are not available in MIMIC-CXR. For consistency, we restrict analysis to the same subsets used in the earlier domain shift experiments (posterior–anterior view only, with the same six harmonised labels). Label prevalences for these six findings across all four datasets are summarised in Table~\ref{tab:classdist}. Demographic distributions (age and sex) for the three datasets with available information are presented in Table~\ref{tab:demographics}.
PadChest and ChestX-ray14 exhibit an almost even gender distribution, while CheXpert shows a notable male majority. Both CheXpert and ChestX-ray14 are dominated by patients between 40 and 65 years, while PadChest is skewed towards older patients, with the majority over 65 years. Patients under 40 represent a minority in both CheXpert and PadChest, and are particularly rare in PadChest, where they make up only 6.3\% of cases.

\begin{table}[h]
\vspace{-5pt}
\caption{Demographic statistics by dataset (subsets from Section 3\ref{data}). No data for MIMIC-CXR.}
\label{tab:demographics}
\begin{tabular}{llllll}
\hline
 & & \textbf{CheXpert} & \textbf{ChestX-ray14} & \textbf{PadChest} \\
\hline
\multirow{3}{*}{Age (\%)} & < 40 & 19.1 & 31.8 & 6.3 \\
& 40 - 65 & 48.6 & 56.3 & 31.1 \\
& > 65 & 32.3 & 11.9 & 62.6 \\
\hline
\multirow{2}{*}{Sex (\%)} & Female & 34.4 & 45.4 & 52.9 \\
& Male & 65.6 & 54.6 & 47.1 \\
\hline
\end{tabular}
\vspace{-5pt}
\end{table}

\vspace{-8pt}
\subsection{Dataset Source Classification}
\begin{figure}[h]
\vspace{-8pt}
\centering
\begin{subfigure}{0.24\textwidth}
    \includegraphics[width=\linewidth]{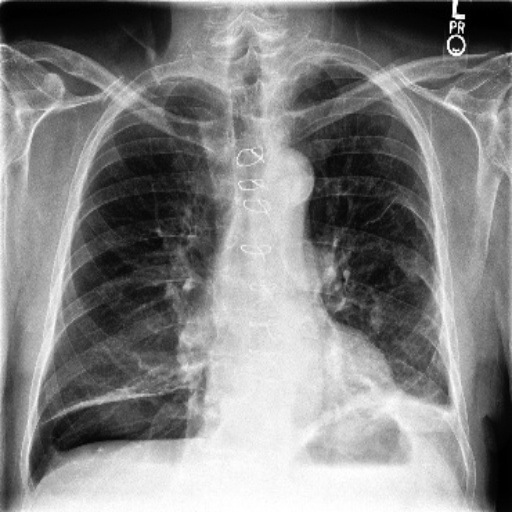}
    \caption{CheXpert}
\end{subfigure}
\begin{subfigure}{0.24\textwidth}
    \includegraphics[width=\linewidth]{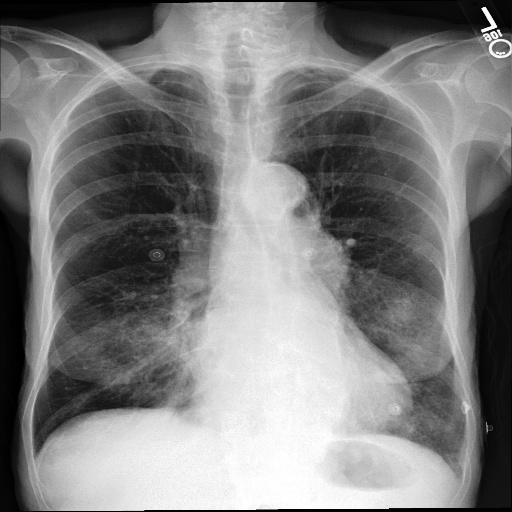}
    \caption{MIMIC-CXR}
\end{subfigure}
\begin{subfigure}{0.24\textwidth}
    \includegraphics[width=\linewidth]{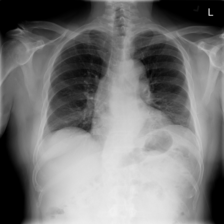}
    \caption{ChestX-ray14}
\end{subfigure}
\begin{subfigure}{0.24\textwidth}
    \includegraphics[width=\linewidth]{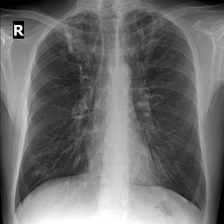}
    \caption{PadChest}
\end{subfigure}
\begin{subfigure}{0.24\textwidth}
    \includegraphics[width=\linewidth]{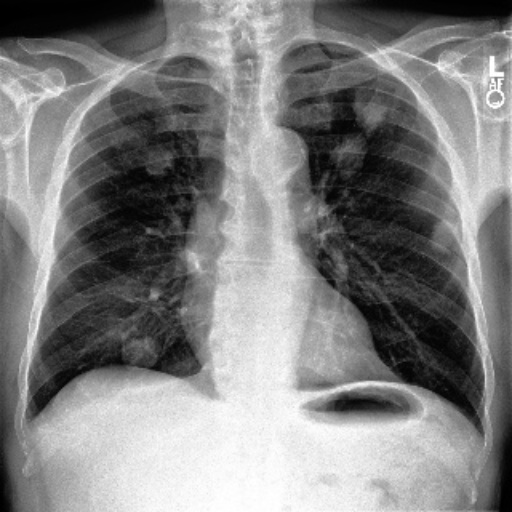}
    \caption{CheXpert}
\end{subfigure}
\begin{subfigure}{0.24\textwidth}
    \includegraphics[width=\linewidth]{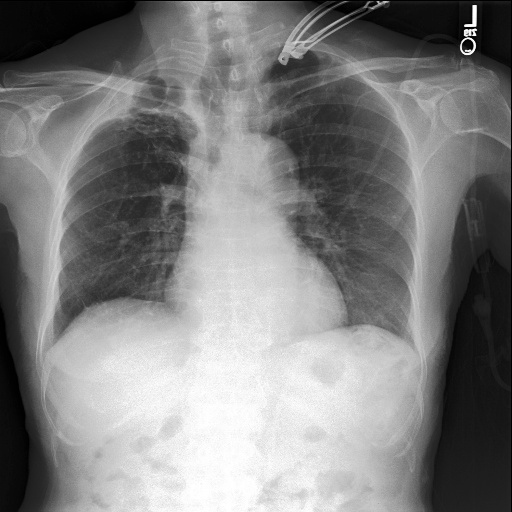}
    \caption{MIMIC-CXR}
\end{subfigure}
\begin{subfigure}{0.24\textwidth}
    \includegraphics[width=\linewidth]{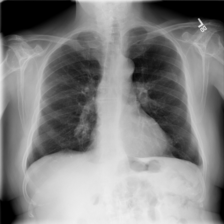}
    \caption{ChestX-ray14}
\end{subfigure}
\begin{subfigure}{0.24\textwidth}
    \includegraphics[width=\linewidth]{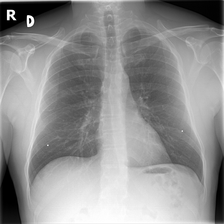}
    \caption{PadChest}
\end{subfigure}
    \caption{Example chest radiographs from MIMIC-CXR, CheXpert, ChestX-ray14 and PadChest. Each image contains text artefacts.}\label{fig:artefacts}
    \vspace{-8pt}
\end{figure}

To assess whether dataset-specific signals are present in images independent of pathology, we train a 4-way classifier to predict dataset source (MIMIC-CXR, CheXpert, ChestX-ray14, PadChest).
We use DenseNet-161, which achieved the highest internal AUROC and AUPRC in three of the four datasets in our domain shift analysis. Using the strongest-performing architecture ensures that any ability to distinguish dataset sources reflects genuine learnable signatures rather than limitations of model capacity. This DenseNet-161 backbone is trained from scratch on our four training datasets, using only pixel data. Performance is evaluated using Sensitivity, Specificity and F1-Score for each of the four classes (source datasets). As chance-level accuracy is 25\%, substantially higher performance would indicate that non-pathological features (e.g., text markers, intensity scaling, acquisition artefacts) allow models to distinguish datasets. Results are shown in Table~\ref{tab:sourceclf}. Example images from each source dataset are shown in Figure \ref{fig:artefacts}.

The DenseNet-161 dataset-source classifier achieved near-perfect performance across all four classes (source datasets), far exceeding the 25\% chance level. This indicates that public chest radiography datasets contain strong, easily learnable site-specific signatures, which are deeply embedded in the data. These signals are unrelated to pathology and instead likely reflect differences in acquisition protocols, scanner hardware, image post-processing pipelines, or embedded text markers. This finding highlights a critical shortcoming of current public benchmarks - models may exploit non-clinical cues that distinguish datasets rather than learning generalisable representations of disease.

Unlike prior work that treats dataset identification primarily as a diagnostic in isolation or as a basis for debiasing objectives, we integrate dataset-source classification with cross-dataset generalisation results and demographic subgroup analyses under fixed clinical operating thresholds. This unified evaluation links institutional signatures, domain shift, and demographic imbalance, allowing us to quantify how dataset bias manifests as external performance degradation and subgroup disparities in widely used public chest radiography benchmarks.

\begin{table}[h]
\vspace{-5pt}
\caption{Performance of the DenseNet-161 dataset-source classifier trained to distinguish the four datasets from images alone. Performance reported as Sensitivity,  Specificity and F1-score for each individual source, as well as on the entire dataset.}
\label{tab:sourceclf}
\begin{tabular}{llllll}
\hline
& \textbf{MIMIC-CXR} & \textbf{CheXpert} & \textbf{ChestX-ray14} & \textbf{PadChest}  & \textbf{Total}  \\
\hline
Sensitivity & 1.000 & 0.897 & 1.000 & 0.999 & 0.985 \\
Specificity & 1.000 & 0.973 & 0.970 & 1.000 & 0.996\\
F1 Score & 1.000 & 0.921 & 0.982 & 1.000 & 0.993 \\
\hline
\end{tabular}
\vspace{-5pt}
\end{table}

\vspace{-8pt}
\subsection{Subgroup Performance Disparities}
We also examine whether model performance differs across demographic subgroups. We focus on the best-performing internal model from the earlier domain shift experiment, DenseNet-161. Subgroups are defined by age (<40, 40–64, $\geq$65) and sex (female, male). 
Using the three datasets with age and sex metadata (CheXpert, ChestX-ray14, PadChest), we compute AUROC, AUPRC, F1, Sensitivity, Specificity, and Balanced Error Rate (BER) per subgroup, using the same fixed thresholds as in the domain shift analysis. BER is a prevalence-independent summary of classification error, computed as $\mathrm{BER}=1-\frac{\mathrm{Sensitivity}+\mathrm{Specificity}}{2}$ from macro-averaged sensitivity and specificity.
Significant differences in performance between subgroups would indicate that models inherit demographic biases from their training data. Table~\ref{tab:subgroup} summarises subgroup-specific results.

When stratified by sex, the model consistently achieved higher performance on chest radiographs belonging to male patients than on those belonging to female patients across all datasets. In CheXpert, this corresponds with the male predominance in the dataset (65.6\%), where the model achieves AUROC 0.853 and F1 0.791 on male scans compared to 0.798 and 0.635 on female scans. A similar but less pronounced pattern is observed in ChestX-ray14 and PadChest, both of which have more balanced sex distributions.  
When stratified by age, the 40–64 cohort consistently achieved the strongest performance, with AUROCs of 0.883, 0.938, and 0.966 in CheXpert, ChestX-ray14, and PadChest, respectively. Patients aged under 40 performed worst across all datasets (F1 0.436, 0.622, and 0.339), reflecting their minority representation (19.1\%, 31.8\%, and 6.3\%). The over 65s cohort showed variable outcomes: in CheXpert and PadChest, where older patients were well represented (32.3\% and 62.6\%), performance was moderate (F1 0.557 and 0.702), whereas in ChestX-ray14, where they made up only 11.9\% of the cohort, specificity collapsed (0.103), reducing F1 to 0.470.  
Model performance aligns closely with the demographic distributions of the training datasets, with underrepresented groups—especially younger and older patients—showing consistently poorer outcomes. These disparities are further reflected by elevated BER, suggesting unequal diagnostic reliability across subpopulations.


\begin{table}[h]
\vspace{-5pt}
\caption{Subgroup-specific performance (AUROC, AUPRC, F1, Sensitivity, Specificity, Balanced Error Rate (BER) for the DenseNet-161 model architecture. Metrics macro averaged over six pathology labels. Results shown for age and sex subgroups, for each dataset containing relevant patient metadata.}
\label{tab:subgroup}
\begin{tabular}{llllllll}
\hline
\textbf{Dataset} & \textbf{Subgroup}  & \textbf{AUROC} & \textbf{AUPRC} & \textbf{F1} & \textbf{Sensitivity} & \textbf{Specificity} & \textbf{BER}\\
\hline
\multirow{5}{*}{CheXpert} & Female & 0.798 & 0.476 & 0.635  &  0.783& 0.570& 0.324 \\
& Male & 0.853 & 0.620 &  0.791 & 0.750  & 0.800 & 0.225\\
& Age < 40  & 0.597 & 0.330 & 0.436  & 0.784 & 0.270& 0.473\\
& Age 40 - 64 & 0.883 & 0.645 &  0.772 & 0.912 &  0.733& 0.178\\
& Age > 65 & 0.746 & 0.461 &  0.557 & 0.792 & 0.488& 0.360\\
\hline
\multirow{5}{*}{CXR14} & Female & 0.901 & 0.707 & 0.702 & 0.812 & 0.680 & 0.254\\
& Male & 0.933 & 0.790 & 0.788 & 0.894 & 0.776& 0.165\\
& Age < 40  & 0.851 & 0.622 & 0.622 & 0.805 & 0.209 & 0.493\\
& Age 40 - 64 & 0.938 & 0.785 & 0.839  &  0.895& 0.724 & 0.191\\
& Age > 65 & 0.612 & 0.503 & 0.470  & 0.921 & 0.103& 0.488\\
\hline
\multirow{5}{*}{PadChest} & Female & 0.913 & 0.832 &  0.751 & 0.792 & 0.680 &0.264\\
& Male & 0.958 & 0.887 &  0.784 & 0.746 & 0.794& 0.230\\
& Age < 40  & 0.682 & 0.377 & 0.339 & 0.772 & 0.289& 0.470\\
& Age 40 - 64 & 0.966 & 0.860 & 0.720  & 0.812 & 0.700 & 0.244\\
& Age > 65 & 0.824 & 0.718 &  0.702 & 0.781 & 0.683& 0.268\\
\hline
\end{tabular}
\vspace{-5pt}
\end{table}

\vspace{-5pt}
\subsection{Summary of Findings}
Our analysis of dataset bias highlights two key vulnerabilities. First, chest radiography datasets contain strong and easily learnable site-specific signatures, enabling standard architectures such as DenseNet-161 to predict dataset source with near-perfect accuracy. These signatures likely arise from differences in acquisition protocols, scanner hardware, or embedded artefacts, and illustrate the risk that models may exploit dataset-specific cues rather than learning generalisable disease features. Second, subgroup analyses demonstrate systematic disparities in model performance across demographic cohorts. Performance is consistently higher on scans from the most represented age and sex subgroups, while minority groups—notably patients under 40 and over 65 years—exhibit substantially poorer results. This indicates that underrepresented cohorts are more vulnerable to degraded model performance. Subgroup disparities are further reflected by prevalence-independent BER, which is consistently higher for underrepresented age and sex cohorts.
Taken together, these findings show that dataset bias arises both from technical acquisition signatures and from imbalanced demographic representation. When considered alongside the domain shift results, they demonstrate that limitations in generalisability stem not only from inter-institutional variation, but also from inherent biases within individual datasets, with important implications for fairness and reliability in clinical deployment.

\vspace{-8pt}
\section{Expert Evaluation: Label Quality}

To assess label quality, we compared the public ground-truth labels of MIMIC-CXR and CheXpert against the independent judgments of two board-certified radiologists. These two datasets were chosen due to their inclusion of free-text radiology reports, and their use of the same automated labeller. Each expert reviewed a stratified subset of 120 images per dataset (20 per class across six classes: \textbf{No Finding, Lung Cancer, Cardiomegaly, Pleural Effusion, Pneumonia,} and \textbf{Pneumothorax}). For every image, experts recorded a binary Agree/Disagree with the dataset label and, when disagreeing, proposed a corrected pathology. We use these datasets because both provide paired radiology reports, enabling a secondary analysis of agreement between expert judgments and original report text in addition to automatically generated labels. The sample size reflects practical time constraints and availability of radiologists.

\vspace{-5pt}
\subsection{Agreement with Dataset Labels}
Per-expert agreement with dataset labels (by class and overall) is summarised in Table \ref{tab:expert}. Across both datasets, overall expert–label agreement is below 60\% for each expert, which is concerningly low. On MIMIC-CXR, \textbf{No Finding} and \textbf{Cardiomegaly} achieve the highest agreement for both experts, whereas \textbf{Lung Cancer, Pneumothorax,} and \textbf{Pneumonia} fall below 50\%. On CheXpert, \textbf{No Finding} and \textbf{Pleural Effusion} show relatively higher agreement, while \textbf{Lung Cancer, Pneumothorax,} and \textbf{Pneumonia} are below 40\% for both experts.

These class-specific patterns are clinically plausible. The weak performance of \textbf{Pneumonia} and \textbf{Pneumothorax} is consistent with the difficulty of detecting subtle opacities and the limited visibility of pneumothorax on frontal chest radiographs. The poor \textbf{Lung Cancer} performance likely reflects limitations of automatic text mining pipelines (e.g., confusion of primary vs metastatic disease, treated lesions, or non-neoplastic masses). In our subset, many \textbf{Lung Cancer} disagreements were re-labelled as \textbf{No Finding} (n=18) or \textbf{Infection / Effusion} (n=13), suggesting over-inclusive use of the “cancer” label.
When experts disagreed with the assigned ground-truth label, they most commonly reassigned \textbf{No Finding} to \textit{Cardiomegaly} or \textit{Pleural Effusion}; \textbf{Lung Cancer} to \textit{No Finding, Effusion, Infection,} or \textit{Atelectasis}; \textbf{Cardiomegaly} to \textit{No Finding} or \textit{Effusion}; \textbf{Pleural Effusion} to \textit{No Finding, Infection, Consolidation,} or \textit{Lung Cancer}; \textbf{Pneumonia} to \textit{No Finding, Effusion, Cardiomegaly,} or \textit{Lung Cancer}; and \textbf{Pneumothorax} to \textit{No Finding, Effusion,} or \textit{Atelectasis}.

There were also cases in which the two experts disagreed with each other (one agreed with the dataset label while the other disagreed): 24/120 (20.0\%) in MIMIC-CXR and 23/120 (19.2\%) in CheXpert. Of these, 19/24 MIMIC-CXR and 21/23 CheXpert expert-disagreement cases were instances where one expert agreed with the assigned pathology label, and the other instead labelled it as \textbf{No Finding}, or said they could not make out the pathology. This implies that these disagreements stem from dataset image quality. Disagreements were most common for visually subtle pathologies \textbf{Pneumonia, Pneumothorax,} and \textbf{Pleural Effusion}—the same classes with the lowest expert–label agreement (Table \ref{tab:expert})—indicating that a substantial component of the apparent label noise reflects genuine reader variability on frontal CXR, not solely dataset mislabelling.

\begin{table}[h!]
\vspace{-8pt}
\caption{Agreement of two board-certified radiologists with public dataset ground truth labels, for a subset of 120 images from MIMIC-CXR and CheXpert. Experts gave binary Agree/Disagree responses.}
\label{tab:expert}
\begin{tabular}{lllll}
\hline
\multirow{2}{*}{} &
\multicolumn{2}{c}{\textbf{MIMIC-CXR}} & \multicolumn{2}{c}{\textbf{CheXpert}}\\
 & \textbf{Expert 1} & \textbf{Expert 2} & \textbf{Expert 1} & \textbf{Expert 2} \\
\hline
No Finding & \textbf{0.85} & \textbf{0.80} & \textbf{0.85} & 0.75 \\
Lung Cancer & 0.15 & 0.30 & \textbf{0.30} & \textbf{0.40} \\
Cardiomegaly & \textbf{0.80} & \textbf{0.95} & 0.50 & 0.65 \\
Pleural Effusion & 0.55 & \textbf{0.70} & \textbf{0.70} & 0.65 \\
Pneumonia & \textbf{0.45} & \textbf{0.40} & 0.35 & 0.15 \\
Pneumothorax & \textbf{0.30} & \textbf{0.40} & 0.20 & \textbf{0.40} \\
\hline
Total & \textbf{0.517} & \textbf{0.591} & 0.493 & 0.500 \\
\hline
\end{tabular}
\vspace{-10pt}
\end{table}

\vspace{-5pt}
\subsection{Agreement with Original Radiology Reports}

Because MIMIC-CXR and CheXpert release anonymised reports alongside images—and their public labels are derived from those reports—we also assessed whether expert judgments align with the original report text. For each image where at least one expert disagreed with the assigned label, we retrieved the corresponding report and reviewed the \textit{Findings} and \textit{Impression} sections. Each case was categorised as \textbf{Expert-Supporting} (report supports expert’s alternate label), \textbf{Dataset-Supporting} (report supports original dataset label), or \textbf{Neither} (report disagrees with both).
In MIMIC-CXR, expert–label disagreements occurred in 58/120 cases for Expert 1 and 49/120 for Expert 2. Considering all images with disagreement from at least one expert gives 71 cases. Of these, 61 were deemed Expert-Supporting, 8 Dataset-Supporting, and 5 Neither. Expert-Supporting cases were especially frequent where experts re-labelled cases to \textbf{No Finding}, underscoring known challenges for text mining with negation handling. Dataset-Supporting cases were mainly where \textbf{Pneumothorax} or \textbf{Pneumonia} had been replaced with \textbf{Pleural Effusion}. In CheXpert, expert–label disagreements occurred in 61/120 (Expert 1) and 60/120 (Expert 2) cases, corresponding to 70 unique images with disagreement from at least one expert. Of these, 64 were Expert-Supporting, 5 Dataset-Supporting, and 1 Neither. Per-class patterns mirrored those of MIMIC-CXR.

\begin{figure}[h!]
    \centering
    \begin{subfigure}{0.7\textwidth}
        \includegraphics[width=\linewidth]{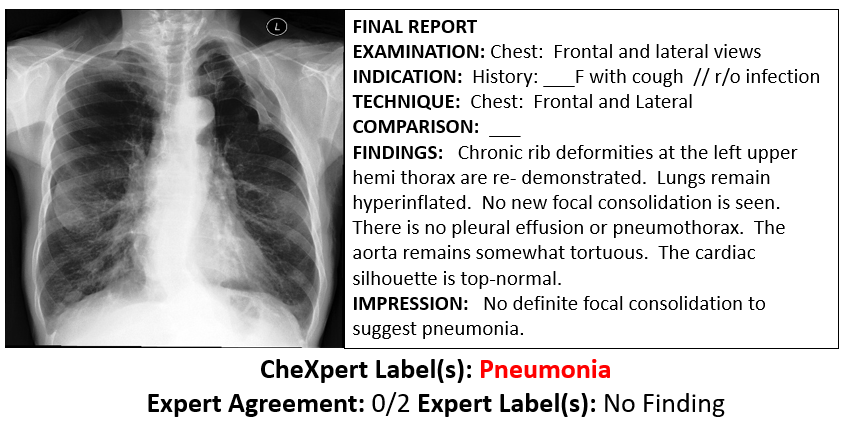}
    \end{subfigure}
    \begin{subfigure}{0.7\textwidth}
        \includegraphics[width=\linewidth]{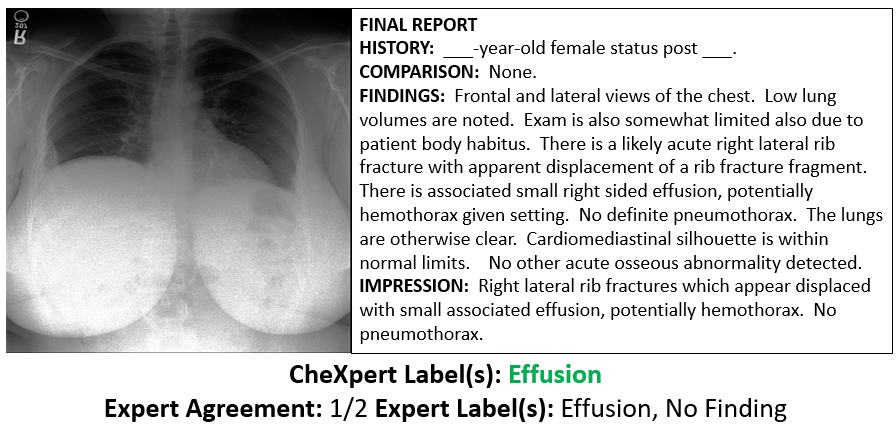}
    \end{subfigure}
    \caption{Example image-report pairs from the MIMIC-CXR dataset, with corresponding labels as extracted by the CheXpert labeller. The first case is labelled Pneumonia, however the report indicates no abnormality. Both experts agree with the report, not the label. The second case is labelled Effusion, with a small effusion described in the report. One expert agreed with the label, and the other found no abnormality.}
    \label{fig:agree}
\end{figure}

Taken together, these findings indicate low expert–label agreement, with a substantial number of discrepancies attributable to report-to-label extraction errors (e.g., misinterpretation of context and negation). Examples are shown in Figure \ref{fig:agree}. This evaluation is limited by a small, stratified sample, two radiologists, and a binary Agree/Disagree protocol that does not capture uncertainty. Future studies should incorporate confidence scales and larger, multi-reader panels to better characterise label quality.

\vspace{-8pt}
\section{Conclusion}

Public chest radiography datasets have enabled rapid progress in medical deep learning, but they also harbour limitations that affect clinical validity—notably institutional bias and imperfect label quality. Our empirical analyses show that models with strong internal performance exhibit substantial degradation on external datasets, particularly in F1 and AUPRC, consistent with calibration drift and shortcut learning under domain shift. These effects were reproducible across diverse architectures, indicating that performance decrease is driven more by the source–target dataset pairing than by model choice. Dataset design, rather than model architecture, is therefore the dominant constraint.

Our dataset bias analysis demonstrates that chest radiography datasets contain strong site-specific acquisition signatures, which models can exploit to distinguish datasets with near-perfect accuracy. Subgroup analyses further reveal systematic disparities in model performance across age and sex cohorts, with consistently higher performance on majority groups and degraded results for underrepresented populations such as younger and older patients. These findings highlight that bias arises not only from inter-institutional differences but also from intrinsic demographic and technical imbalances within individual datasets, raising concerns about fairness and equity in clinical deployment.  

Our expert audit further highlights deficiencies in automatically generated pathology labels. Overall expert–label agreement was <60\% for two independent radiologists, with the weakest agreement for Pneumonia, Pneumothorax, and Lung Cancer. In contrast, expert assessments aligned far more closely with the original radiology reports, implicating report-to-label extraction errors—especially misinterpretation of context and negation—as a major source of noise. Therefore, current automated labels are adequate for scale but inadequate as a clinical reference standard, helping explain why models that excel internally often fail to generalise.

Improving the clinical reliability of models trained and evaluated on these datasets requires (i) larger clinician-validated subsets and report-aware relabelling of the underlying corpora; (ii) transparent documentation of labelling pipelines and explicit handling of uncertainty; and (iii) multi-centre, demographically diverse datasets with systematic artefact auditing to mitigate shortcut learning and subgroup disparities. Together, these measures shift the field from benchmark optimisation toward trustworthy deployment, in which performance reflects real-world diagnostic reasoning across diverse patient populations and clinical settings.

\paragraph{Data Availability.} This study uses four publicly available chest radiography datasets: MIMIC-CXR, CheXpert, ChestX-ray14, and PadChest. MIMIC-CXR and CheXpert are accessible via PhysioNet following completion of the required data use agreements and credentialing. ChestX-ray14 and PadChest are openly available from their respective project websites. All experiments were implemented in PyTorch using standard, publicly available model architectures. Full details of model architectures, training procedures and evaluation protocols are provided in the manuscript to support reproducibility. 

\ack{We thank Dr Rishi Ramaesh and Dr Allan Green from NHS Lothian for providing their clinical expertise in both experiment design and assessing dataset label agreement.}


\bibliographystyle{vancouver}
\bibliography{sample}

\end{document}